\documentclass[11pt]{article}

\usepackage[preprint]{acl}

\usepackage{times}
\usepackage{latexsym}

\usepackage[T1]{fontenc}

\usepackage[utf8]{inputenc}
\usepackage{pifont}
\DeclareUnicodeCharacter{270D}{\ding{45}}
\DeclareUnicodeCharacter{FE0F}{}

\usepackage[stretch=30,shrink=30]{microtype}
\usepackage{hyphenat}
\usepackage{xurl}
\usepackage{etoolbox}

\AtBeginDocument{\fussy\tolerance=1000\emergencystretch=1em}
\apptocmd{\thebibliography}{\fussy\tolerance=1000\emergencystretch=1em}{}{
  \PackageError{paper}{Could not adjust bibliography spacing}{}
}

\usepackage{inconsolata}

\usepackage{graphicx}

\usepackage{graphicx}

\usepackage{xspace}
\usepackage{caption}
\usepackage{wrapfig}
\usepackage[percent]{overpic}
\usepackage{float}
\usepackage{booktabs} 
\usepackage{collcell}
\newcommand{\mytexttt}[1]{\raggedright\texttt{#1}}
\newcolumntype{M}[1]{>{\collectcell\mytexttt}p{#1}<{\endcollectcell}}
\usepackage{inconsolata}
\usepackage[most]{tcolorbox}
\usepackage{listings}
\usepackage{fontawesome5} 
\usepackage[svgnames]{xcolor}
\usepackage{amssymb}
\usepackage{multirow}

\usepackage{hyperref}
\usepackage{cleveref}

\crefformat{section}{\S#2#1#3}
\Crefformat{section}{\S#2#1#3}

\usepackage{subcaption}

\usepackage{xcolor}
\usepackage{longtable}

\usepackage{makecell}

\usepackage{xcolor}
\definecolor{increase}{RGB}{0,140,70}
\definecolor{decrease}{RGB}{200,40,40}

\newcommand{\fakeparagraph}[1]{\vspace{1mm}\noindent\textbf{#1}}

\newif\ifshowcomment
\showcommenttrue   

\usepackage[most]{tcolorbox}

\newtcolorbox{propositionbox}[1][]{
  colback=gray!8,        
  colframe=gray!40,      
  boxrule=0.4pt,
  arc=2pt,
  left=8pt, right=8pt, top=6pt, bottom=6pt,
  #1
}

\usepackage{longtable}
\usepackage{booktabs}
\usepackage{array}
\usepackage{xcolor}
\usepackage{colortbl}

\usepackage{amsmath}

\title{Evaluating LLM Creativity through Long-Tail Performance}

\author{
Yichen Wang \\
University of Chicago \\
{\small\texttt{yichenzw@uchicago.edu}} \\\And
Tenghao Huang \\
Microsoft \\
{\small\texttt{tenghaohuang@microsoft.com}} \\\And
Vishakh Padmakumar \\
Stanford University \\
{\small\texttt{vishakhp@stanford.edu}} \\\And
Mina Lee \\
University of Chicago \\
{\small\texttt{mnlee@uchicago.edu}} \\
}

\begin{document}
\maketitle
\begin{abstract}
Creativity benchmarks for large language models (LLMs) vary widely in task and metric design.
This raises the question of whether they capture a common notion of creativity.
We propose that this shared target can be formalized as task-specific performance in the long tail of a reference input-output distribution, which describes how common input-output pairs are relative to a chosen source of text.
We ground this proposition in the novelty-quality product formulation and evaluate it empirically across 15 benchmarks and 26 open-weight models, using n-gram absence from the pretraining data as a novelty proxy.
Our results show that creativity benchmarks lie further in the tail regions than standard benchmarks. 
Moreover, on both creativity and standard benchmarks, model rankings from higher-tail subsets of each benchmark align more closely with human creativity judgments in LM-Arena.
Building on these findings, we introduce q@Tail($\alpha$), a lower-cost creativity diagnostic that measures mean task quality on the tail fraction of an existing benchmark.
Applying it to creativity-promoting methods reveals substantial room to improve LLM creativity through stronger tail performance.\footnote{Code and data are available at \texttt{\url{https://github.com/YichenZW/longtail-creativity}}. Taxonomy and reading list are at \texttt{\url{https://github.com/YichenZW/awesome_llm_creativity}}.}

\end{abstract}


\begin{figure}[!t]
\centering
\includegraphics[width=\linewidth]{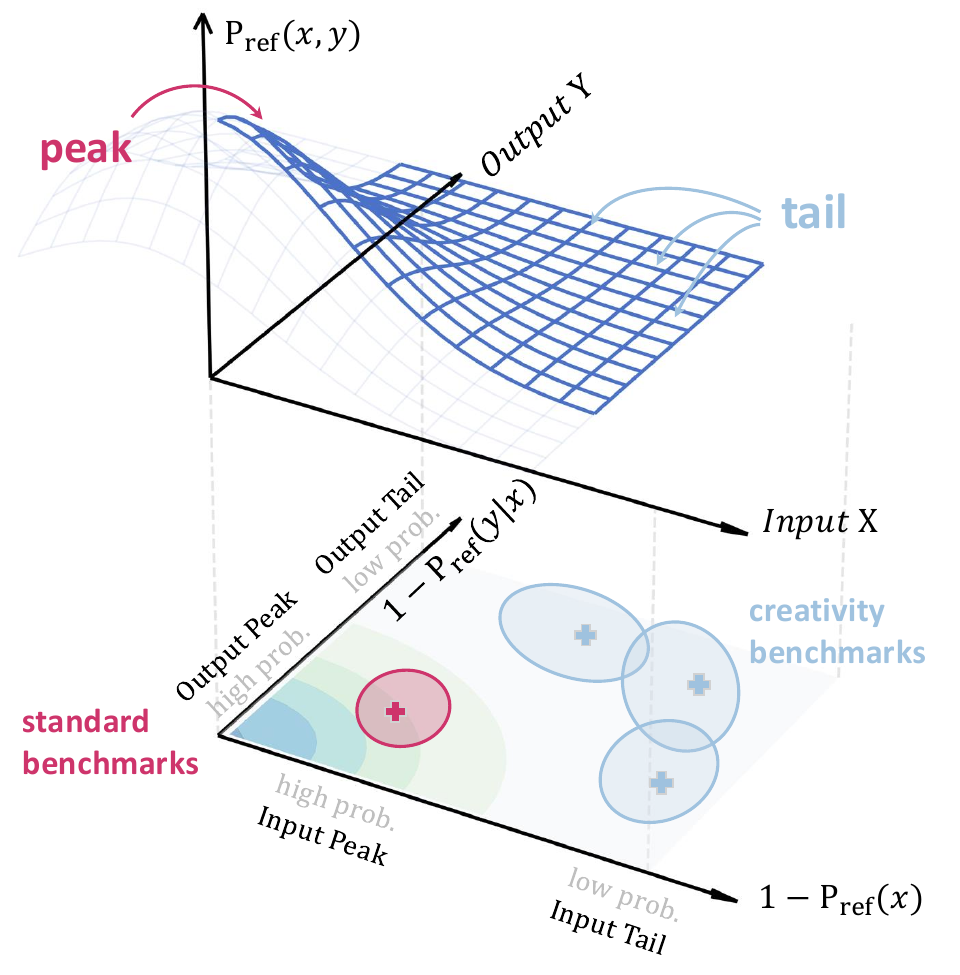}
\caption{\textbf{
Creativity benchmarks lie in the long tail of a reference input-output distribution.
} 
{\emph{Top:}} 
A reference distribution $P_{\text{ref}}(x, y)$ assigns higher probability to common input-output pairs and lower probability to uncommon or novel ones. The surface is schematic because possible inputs and outputs have no natural ordering.
{\emph{Bottom:}} 
The same space is re-parameterized by input improbability, $1 - P_{\text{ref}}(x)$, and output improbability, $1 - P_{\text{ref}}(y \mid x)$. 
Standard benchmarks cluster in high-probability regions (i.e., \emph{Peak}), where inputs and expected outputs are probable and common. 
Creativity benchmarks shift evaluation toward low-probability regions (i.e., \emph{Tail}), where the input is improbable, the output is improbable given the input, or both.
The figure is illusive but depicted peak-tail divergence is empirically corroborated in \Cref{sec:exp1}.
}
\label{fig:intro}
\end{figure}

\section{Introduction}

Creativity is a central capability of intelligence, with a long history of study in human cognition \citep{stein1953creativity, barron1955disposition, amabile1983social, boden1994creative, runco2012standard} 
and growing attention in artificial intelligence.
As large language models (LLMs) are increasingly used for open-ended tasks such as writing, ideation, planning, problem solving, and discovery \citep{chakrabarty2024artartificelargelanguage,ruan2025liveideabenchevaluatingllmsdivergent,tian2025macgyverlargelanguagemodels,gottweis2026accelerating},
creativity has become an important target for model evaluation. 
Yet despite decades of study, human creativity lacks a consensus definition across psychology, cognitive science, design, education, and computational creativity (\Cref{sec:rel-theory}).

Existing LLM creativity benchmarks differ in both task and metric design.
Task dimensions include the objective (such as writing, problem solving, or ideation) and the domain (such as language, mathematics, or science).
Metric dimensions include the granularity at which outputs are evaluated (such as n-grams, sentences, or paragraphs). We organize these in detail in a taxonomy in \Cref{sec:taxonomy}.
This variety raises the question of whether such heterogeneous benchmarks nevertheless capture a common notion of creativity.
If they do, this shared target is not obvious, given the many dimensions along which these benchmarks differ. 
Identifying it would provide a principled basis for designing, validating, and comparing LLM creativity evaluations.

In this work, we propose that this shared target can be formalized as performance in the long tail of a reference input-output distribution (\Cref{sec:proposition}). 
Here, a reference distribution specifies how common inputs and outputs are relative to a chosen source of text, such as a pretraining corpus.
Under this view, one relevant aspect of evaluating LLM creativity is asking whether a model can produce high-quality outputs in low-probability regions of that distribution (\Cref{fig:intro}).
Standard benchmarks tend to evaluate model performance in high-probability regions (i.e., peak) of this distribution, where inputs and expected outputs are probable and common. 
Creativity benchmarks, by contrast, shift evaluation toward the long tail: regions where the input is improbable, the output is improbable given the input, or the input-output pair is otherwise improbable under the reference distribution.


We investigate this proposition through a literature review (\Cref{sec:rel-theory};\Cref{sec:taxonomy}), theoretical analysis (\cref{sec:theorm}), and empirical evaluation (\cref{sec:exp1}-\cref{sec:exp3}).
First, we survey 82 papers and organize their creativity evaluation designs into a taxonomy (\Cref{sec:rel-theory};\Cref{sec:taxonomy}).
Next, we adopt a novelty-quality product formulation of creativity \citep{padmakumar2025measuringllmnoveltyfrontier}.
Under this formulation, the same quality difference contributes more to the benchmark-level creativity score when it occurs in higher-novelty regions (\cref{sec:theorm}).
Finally, our empirical analysis shows that creativity benchmarks lie further in the tail of the reference distribution than standard benchmarks (\Cref{sec:exp1}).
And in practice, common creativity elicitation operations push benchmarks toward distinct tail regions (\Cref{sec:exp1_explain}).
On both creativity and standard benchmarks, rankings from higher-tail subsets agree more closely with the LM Arena creative-writing ranking \citep{chiang2024chatbotarenaopenplatform} than rankings from lower-tail subsets (\Cref{sec:tail-within}).
Building on these findings, we introduce $q@\mathrm{Tail}(\alpha)$, a creativity diagnostic that measures mean task quality on the most novel $\alpha$ fraction of an existing benchmark.
We then apply it to a range of test-time and training-time creativity-promoting methods (\Cref{sec:diagnose}). 
Most methods increase output novelty without consistently improving quality in the tail, revealing substantial room to improve LLM creativity through stronger tail performance.

In short, we make three contributions: 

(\textit{i}) we survey LLM creativity evaluations and characterize their design variation; 

(\textit{ii}) we propose long-tail performance under a reference input-output distribution as a shared target and provide theoretical and empirical support;

(\textit{iii}) we introduce a lower-cost tail-performance metric as a creativity diagnostic for existing benchmarks, and assess creativity-promoting methods.

\section{What is Missing in Literature?}
\label{sec:rel-theory}

Despite decades of study in psychology and cognitive science, creativity has no broadly accepted definition \citep{runco2012standard}. The most widely shared definition characterizes creativity as the joint presence of \emph{novelty} and \emph{quality}: 
an output must be uncommon relative to a reference, and also useful, fitting, or valuable \citep{stein1953creativity,barron1955disposition,boden1994creative,amabile1983social}.
This lack of consensus is reflected in LLM creativity benchmarks, which span writing \citep{chakrabarty2024artartificelargelanguage,atmakuru2024cs4,west2025basemodelsbeataligned}, problem solving \citep{ye2025assessing,han2025creativity,tian2025macgyverlargelanguagemodels}, and ideation \citep{chen2023probingcreativitylargelanguage,qiu-hu-2025-deep,ruan2025liveideabenchevaluatingllmsdivergent}. 

%
Existing LLM creativity evaluations differ substantially in task and metric design and can yield contrasting conclusions.
For example, studies using semantic metrics report strong creative performance by LLMs, whereas evaluations based on expert rubrics identify substantial limitations \citep{chen2023probingcreativitylargelanguage,chakrabarty2024artartificelargelanguage}.
Similarly, studies measuring novelty relative to web text report benefits from model scaling, whereas studies measuring diversity across generated responses report declines \citep{padmakumar2025measuringllmnoveltyfrontier,zhang2025noveltybenchevaluatinglanguagemodels}.
Such contrasting findings motivate making evaluation settings explicit when comparing claims about LLM creativity.
Our review of 82 papers organizes task and metric design choices into a nine-dimensional taxonomy (\Cref{sec:taxonomy}).
By making differences in the dimensions, the taxonomy highlights potential sources of disagreement and provides a perspective that may help explain divergent findings.
Prior comparisons and surveys document this fragmentation \citep{lu2025rethinking,hou2025creativityprism,jiang2025artificialhivemindopenendedhomogeneity,ismayilzada2024creativity,franceschelli2024creativity}.

We then move beyond characterizing these differences to a more fundamental challenge: determining whether a common evaluative target underlies these heterogeneous task and metric designs.
Identifying such a target would provide a principled basis for interpreting existing creativity benchmarks and guiding the design of new evaluations.
On the metric side, \citet{padmakumar2025measuringllmnoveltyfrontier} operationalize creativity through novelty and quality, which can explain different metric designs in the literature. 
On the task side, however, the implicit target underlying tasks designed to elicit creativity remains unexplained. 
We propose that creativity benchmark design implicitly shifts evaluation toward task performance in the long tail of a reference input-output distribution (\Cref{sec:proposition}).
Empirically, we demonstrate a shift toward higher novelty in the evaluated creativity benchmarks relative to standard benchmarks (\Cref{sec:exp1}), and examine whether tail performance is positively associated with human judgments of model creativity (\Cref{sec:exp3}).

\section{Evaluate LLM Creativity through Long-Tail Performance}
\label{sec:proposition}

\subsection{Definition}
\label{sec:tail-definition}

Let $\mathcal{X}$ denote the input space (e.g., prompts, problems, or questions) and $\mathcal{Y}$ the output space (e.g., responses, solutions, or answers). 
A \textbf{reference distribution} $P_{\mathrm{ref}}(x,y)$ specifies how common an input-output pair $(x,y)\in\mathcal{X}\times\mathcal{Y}$ is relative to a chosen source of text (\Cref{fig:intro}, \emph{Top}).
The choice of reference $P_{\mathrm{ref}}$ is an evaluation parameter: different sources, such as a pretraining corpus, a human-authored corpus, or a model's self-samples, induce different notions of novelty (discussed in \S\ref{sec:taxonomy}).



The factorization, $P_\text{ref}(x, y) = P_\text{ref}(x) \cdot P_\text{ref}(y \mid x)$, admits a natural 2D parameterization with axes $P_\text{ref}(x)$ and $P_\text{ref}(y \mid x)$ (\Cref{fig:intro}, \emph{Bottom}). Each $(x, y)$ pair is a point in this plane; a benchmark is a finite set of such points, yielding an empirical distribution in this space.
This factorization decomposes the input-output space into four quadrants: the \textit{peak}, where both the input and output are common; the \emph{input-tail}, where the input is novel; the \emph{output-tail}, where the output is novel given the input; and the \emph{joint-tail}, where both are novel. 

Let $s_{\mathrm{ref}}(x,y)
=
\phi\!\left(
P_{\mathrm{ref}}(x)
P_{\mathrm{ref}}(y\mid x)
\right)
\in[0,1]
\label{eq:pair-novelty}$
denote a normalized, monotonically decreasing transformation of the joint reference probability to a \textbf{novelty} score. 
Larger novelty $s_{\mathrm{ref}}(x,y)$ therefore indicates that the input-output pair lies further in the reference tail.
For benchmark $B$, let $\mu_B$ denote the empirical distribution over its evaluated input-output pairs, and let $\mu_B^X$ denote its marginal distribution over inputs.
We define the prompt-level novelty score
$S_B(x)
=
\operatorname{Agg}_{Y\sim\mu_B(\cdot\mid x)}
\left[
s_{\mathrm{ref}}(x,Y)
\right],
\label{eq:benchmark-novelty}$
where the aggregation is fixed across benchmarks and models and maps scores in $[0,1]$ to $[0,1]$.
Thus, $S_B(x)$ summarizes how far the input and its associated outputs (from evaluated models) jointly lie in the reference tail.

\subsection{Theoretical connection}
\label{sec:theorm}

Creativity is commonly characterized as the joint presence of novelty and quality \citep{runco2012standard,boden1994creative,amabile1983social}.
We adopt the novelty-quality product formulation used by \citet{padmakumar2025measuringllmnoveltyfrontier}:
\begin{equation}
\mathrm{Creativity}(x,y)
=
\mathrm{Novelty}(x,y)
\cdot
\mathrm{Quality}(x,y).
\end{equation}

Let $Y_\pi\sim\pi(\cdot\mid X)$ be the output generated by
model $\pi$, and let
$q:\mathcal{X}\times\mathcal{Y}\rightarrow[0,1]$ be a
task-specific quality function. We define the expected quality
of $\pi$ on input $x$ as
$\bar q_\pi(x)
=
\mathbb{E}_{Y_\pi\sim\pi(\cdot\mid x)}
\left[q(x,Y_\pi)\right].$
Under the novelty-quality product formulation, the
benchmark-level creativity score of $\pi$ is
\begin{equation}
C_{\pi,B}
=
\mathbb{E}_{X\sim\mu_B^X}
\left[
S_B(X)\bar q_\pi(X)
\right].
\label{eq:overall-creativity}
\end{equation}

Since $S_B(X)\in[0,1]$, Tonelli's theorem gives the exact identity
\begin{equation}
C_{\pi,B}
=
\int_0^1
H_B(t)A_{\pi,B}(t)\,dt,
\label{eq:layer-cake}
\end{equation}
where $H_B(t)=\Pr_{X\sim\mu_B^X}(S_B(X)\ge t)$ is the mass of the tail at threshold $t$, and
$A_{\pi,B}(t) = \mathbb{E}_{X\sim\mu_B^X} \left[
\bar q_\pi(X)\mid S_B(X)\ge t
\right]$
is the model's quality on that tail.
Thus, Equation~\ref{eq:layer-cake} expresses overall creativity as an aggregation of quality across all nested novelty regions.

Benchmarking primarily concerns ranking with relative model performance.
For a baseline model $\pi_0$, one benchmark gives
\begin{equation}
C_{\pi,B}-C_{\pi_0,B}
=
\int_0^1
H_B(t)
\left[
A_{\pi,B}(t)-A_{\pi_0,B}(t)
\right]dt.
\label{eq:relative-creativity}
\end{equation}
A quality advantage on an input with novelty $S_B(x)$ appears at every threshold $t\le S_B(x)$ and therefore receives total weight $S_B(x)$. 
Consequently, the same quality difference contributes more to the overall creativity difference when it occurs in higher-novelty regions. 

A finite benchmark cannot estimate the complete tail-quality
curve at arbitrarily fine resolution. We therefore use
$q@\mathrm{Tail}_{\pi,B}(\alpha)
=
A_{\pi,B}(Q_{B,\alpha})
\label{eq:q-at-tail}$, $Q_{B,\alpha}$ as the threshold of top-$\alpha$ subset of benchmark $B$,
as a coarse-grained summary of quality in a selected tail
region. It approximates the relative ranking induced by
$C_{\pi,B}$ when differences in the selected tail dominate,
or are ranking-consistent with, the remaining contributions.
We verify this condition empirically in \S\ref{sec:exp3}. See \Cref{app:q-tail-interpretation} for the interpretation of this shared-weight diagnostic.

\section{Experimental Setup}
\label{sec:settings}

This section instantiates \S\ref{sec:proposition} empirically: we specify a concrete reference distribution $P_\text{ref}$, the novelty metric as axes, the model and benchmark sets, and the performance metric used in \S\ref{sec:exp1}-\S\ref{sec:exp3}.

\paragraph{Reference distribution.}
We use the Pile pretraining corpus~\citep{gao2020pile800gbdatasetdiverse} as the reference text source and query $n$-gram occurrences with Infini-Gram-Mini~\citep{liu2025infinigramscalingunboundedngram,xu2026infinigramminiexactngram}.
These queries determine whether an $n$-gram appears in the corpus and supply the corpus-absence novelty scores defined below.
The Pile predates all evaluated benchmarks, mitigating direct contamination that would inflate the measurement.
The choice of $P_{\mathrm{ref}}$ is an evaluation parameter (\Cref{sec:proposition}) and our taxonomy catalogs alternative reference sources (\Cref{sec:taxonomy}). 

\paragraph{Novelty metric.}
We operationalize input and output tailness using $n$-gram novelty, a computationally tractable measure at the scale of the Pile that follows lexical matching approaches used in memorization analysis and retrieval~\citep{merrill-etal-2024-evaluating,liu2025infinigramscalingunboundedngram}.\footnote{Perplexity from a neural language model trained on the Pile and our $n$-gram novelty score, implemented through a corpus-backed statistical gram language model, are both grounded in the same reference corpus.
However, perplexity also depends on how well the model learns different parts of that corpus, potentially introducing model-dependent variation into tailness estimates.
Our implementation queries corpus $n$-gram occurrence statistics directly through Infini-Gram-Mini, avoiding this additional dependence on neural-model fitting. \Cref{app:perplexity-check} reports our check using base-model perplexity.}
For a text $t$, $\text{novelty}_n(t)$ is the fraction of its $n$-grams absent from the corpus; higher $\text{novelty}_n$ indicates higher improbability and novelty, placing $t$ further into the tail region $\mathcal{T}$.
The results in \Cref{sec:exp1,sec:exp3} and their corresponding appendix analyses use $n=4$.
We also smoke test $n=5,6$ and their findings are consistent.
For each input $x$, we first average output novelty across the evaluated models, then take the geometric mean of input novelty and this average to obtain the shared score $S_B(x)$.
Input-output pairs sharing the same input are selected together using $S_B(x)$, ensuring that all models are compared on the same prompts.

\paragraph{Models.}
We select recent model releases to cover contemporary LLMs and use open-weight models to support reproducible evaluation.
We include a range of model sizes within our available computational resources to examine performance across model scales.
Most selected models are covered by the LM Arena creative-writing leaderboard, providing an external human-preference reference for the ranking analysis (\Cref{sec:exp3}).
The resulting 26 models are drawn from the Llama~\citep{grattafiori2024llama3},
Qwen~\citep{yang2025qwen3,qwen2026qwen35}, 
OLMo~\citep{team2025olmo3}, 
Gemma~\citep{kamath2025gemma3}, 
and Mistral~\citep{mistral2025small32} families, 
plus GPT-OSS-20B~\citep{openai2025gptoss}, 
Phi-4~\citep{abdin2024phi4}, 
Nanbeige4-3B-Thinking~\citep{yang2025nanbeige4}, 
and Nemotron-3-Nano-30B~\citep{nvidia2025nemotron3nano} 
(detailed in \Cref{tab:models-creative-writing}).
The set spans both instruction-tuned and reasoning-tuned models; we enable thinking mode by default for the latter and use each model's recommended decoding configuration.

\begin{figure*}[t]
    \centering
    \begin{subfigure}[t]{0.39\textwidth}
        \centering
        \includegraphics[width=\linewidth]{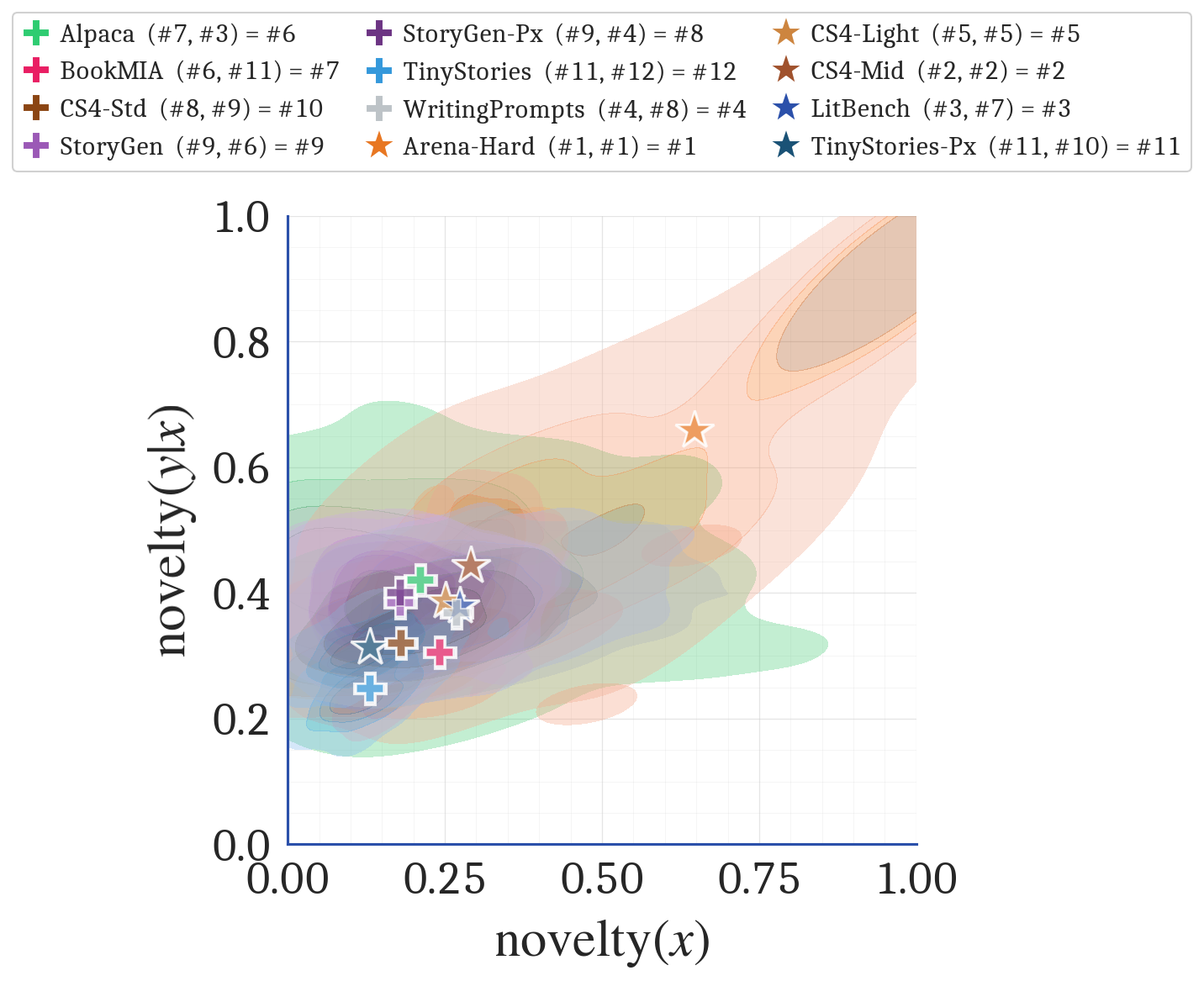}
        \caption{Free-form writing}
        \label{fig:tail-writing}
    \end{subfigure}
    \hfill
    \begin{subfigure}[t]{0.27\textwidth}
        \centering
        \includegraphics[width=\linewidth]{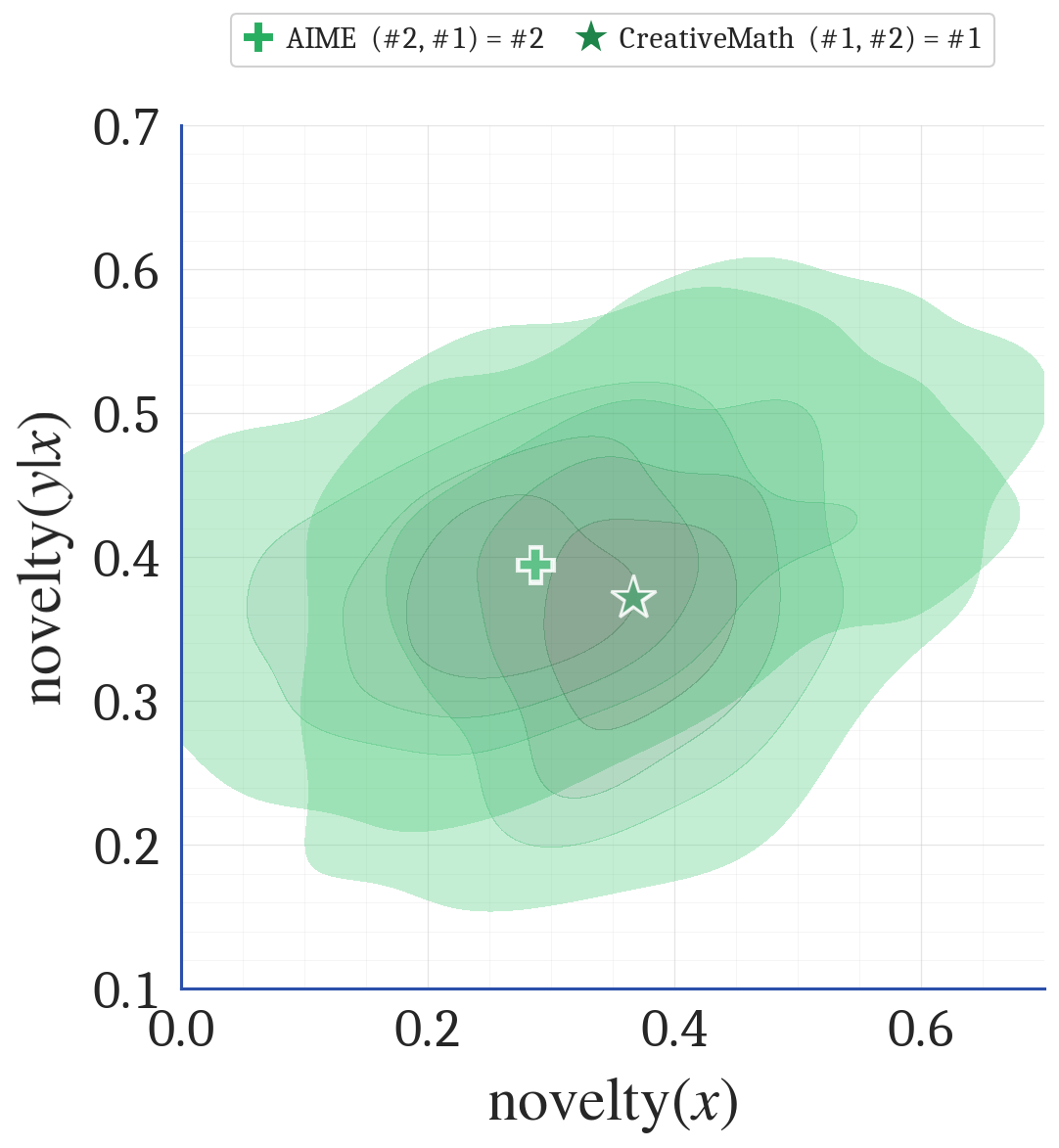}
        \caption{Mathematics reasoning}
        \label{fig:tail-math}
    \end{subfigure}
    \hfill
    \begin{subfigure}[t]{0.32\textwidth}
        \centering
        \includegraphics[width=\linewidth]{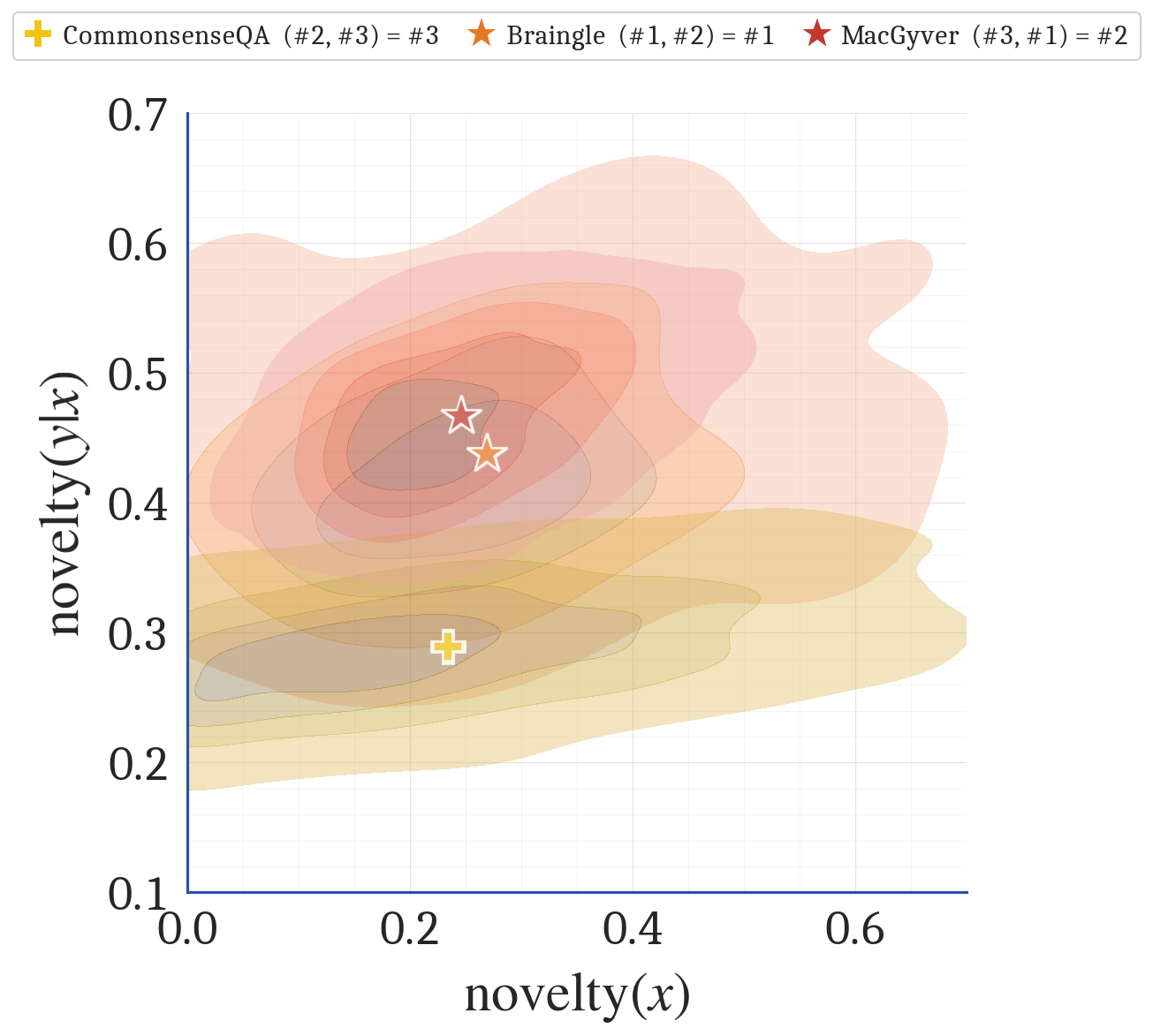}
        \caption{Free-form QA}
        \label{fig:tail-qa}
    \end{subfigure}
    \caption{\textbf{Creativity benchmarks shift toward higher input and/or output novelty than standard benchmarks across all three task domains.}
    2D kernel density estimates of each benchmark in the
    $(\mathrm{novelty}(x),\,\mathrm{novelty}(y\mid x))$ plane.
    Novelty is the fraction of 4-grams absent from the Pile, computed using
    Infini-Gram-Mini (\S\ref{sec:settings}).
    Per-prompt output novelty is averaged across all evaluated models.
    Markers indicate per-benchmark means.}
    \label{fig:tail-by-domain}
\end{figure*}

\paragraph{Benchmarks.}
We select benchmarks using two criteria: creativity benchmarks must have been introduced or used to evaluate LLM creativity in published work, and standard capability benchmarks must cover the same task domains.
This selection allows us to compare the novelty distributions of creativity and standard benchmarks within each task domain.
We evaluate 15 benchmarks across three task domains: free-form writing (10), mathematics reasoning (2), and free-form QA (3).
Creativity benchmarks include Arena-Hard Creative Writing~\citep{li2024arenahard},
CS4~\citep{atmakuru2024cs4},
CreativeMath~\citep{ye2025assessing},
Braingle~\citep{han2025creativity},
and MacGyver~\citep{tian-etal-2024-macgyver};
standard capability benchmarks include Alpaca~\citep{taori2023alpaca},
AIME-All~\citep{balunovic_srimatharena_2025},
and Commonsense-QA~\citep{talmor2019commonsenseqaquestionansweringchallenge}.
The full catalog is provided in \Cref{tab:benchmarks-revised}.

\paragraph{Performance evaluation.}
Even within a single domain, prior creativity benchmarks use heterogeneous performance metrics, e.g., rubric-based LLM-as-judge~\citep{chakrabarty2024art, paech2025eqbench}, reward models~\citep{zhang2025noveltybench}, and reference-similarity scoring~\citep{lu2025salieri} are all common for writing alone (\S\ref{sec:taxonomy} catalogues the broader space). 
To prevent metric choice from confounding tail-vs-peak comparisons, we hold the performance metric fixed \emph{within} each task domain; we do not attempt to compare absolute quality across domains. 
For tasks with ground-truth answers (math, QA), we use exact-match on the extracted final answer via LM-Eval-Harness~\citep{eval-harness}. For open-ended generation (e.g., writing), we use the normalized reward from Skywork-Reward-V2-Llama-3.1-8B~\citep{liu2025skywork}, the current top-ranked model on RewardBench-v2~\citep{malik2025rewardbench2advancingreward}.

\paragraph{Ranking.}
We run the evaluation pipeline for each benchmark $B$ and obtain a model
ranking $r_B$ by sorting all models according to their mean quality score $q$.
As an external reference for human creativity judgment, we use the LM Arena
creative-writing ranking~\citep{chiang2024chatbotarenaopenplatform},\footnote{\url{https://arena.ai/leaderboard/text/creative-writing}.}
which aggregates pairwise human preferences over prompts in the creative-writing
category; we denote this reference ranking as $r_{\mathrm{Arena}}$. In
\Cref{sec:exp3}, we compute Kendall's $\tau(B) := \tau(r_B, r_{\mathrm{Arena}})$
to quantify how closely each benchmark's model ranking aligns with this
human-preference reference.

\section{Creativity Benchmarks Lie in the Tail of $P_{\text{ref}}$}
\label{sec:exp1}

We test whether creativity benchmarks occupy lower-probability regions of $P_{\mathrm{ref}}$ than standard benchmarks from the same task domains.
\Cref{sec:exp1_explain} examines how creativity elicitation operations relate to input-tail, output-tail, and joint-tail regions.

\fakeparagraph{Setup.}
For each benchmark, we plot a two\hyp{}dimensional kernel density estimate in the $(\text{novelty}(x),\allowbreak \text{novelty}(y\mid x))$ plane, using the output-novelty aggregation defined in \Cref{sec:settings}.

\fakeparagraph{Results.}
\textbf{Creativity benchmarks lie further in the tail than standard benchmarks}, shifting toward higher input and/or output novelty. The pattern holds in all three task domains with multiple benchmarks (Figure~\ref{fig:tail-by-domain}A-C). In writing (Figure~\ref{fig:tail-by-domain}A), Arena-Hard Creative Writing sits at the high end of both axes (both approximately $0.6$ to $0.7$), well above standard benchmarks (Alpaca, TinyStories, BookMIA); LitBench and CS4 likewise lie further in the tail than every standard benchmark in this task domain. In math reasoning (Figure~\ref{fig:tail-by-domain}B), CreativeMath lies to the right of AIME-All along $\text{novelty}(x)$. In free-form QA (Figure~\ref{fig:tail-by-domain}C), MacGyver and Braingle lie above CommonsenseQA along $\text{novelty}(y\mid x)$, which anchors the low-novelty end. The pattern is stable across $n\in\{4,5,6\}$ and across model families.
Pooled univariate and multivariate tests confirm that the creativity and standard distributions differ significantly (all $p<0.001$; \Cref{app:distribution-tests}) in the writing domain. 

\subsection{Three elicitation operations, three tail sub-regions}
\label{sec:exp1_explain}

\begin{figure}[t]
    \centering
    \includegraphics[width=0.95\linewidth]{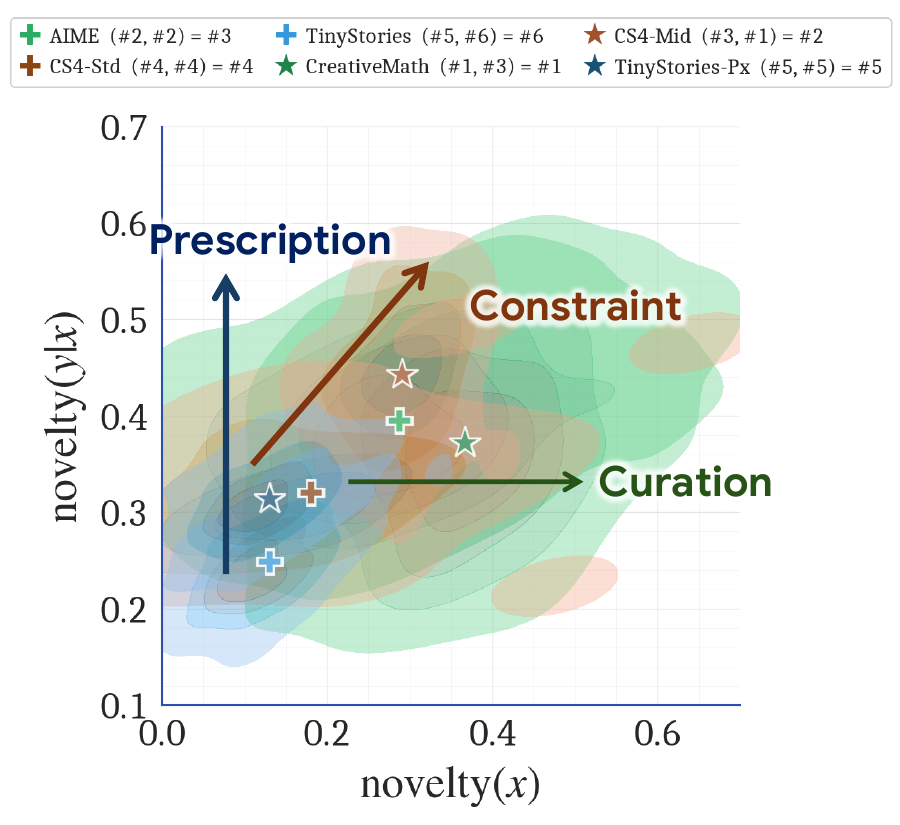}
    \caption{\textbf{Three creativity elicitation operations illustrate shifts toward different tail regions.}
    Aggregated kernel density estimates for creativity and standard benchmarks in the
    $(\mathrm{novelty}(x),\,\mathrm{novelty}(y\mid x))$ plane.
    Arrows illustrate shifts associated with \emph{Curation} toward higher input novelty
    (input-tail), \emph{Prescription} toward higher output novelty (output-tail),
    and \emph{Constraint} toward higher novelty on both axes (joint-tail).
    Operations are defined in \Cref{sec:exp1_explain}; the taxonomy is provided in \Cref{sec:taxonomy}.}
    \label{fig:elicit-subregions}
\end{figure}

The observed shift toward tail regions supports long-tail task performance as one perspective on existing creativity evaluations.
\Cref{fig:elicit-subregions} relates the three creativity elicitation operations in our taxonomy (\Cref{sec:taxonomy}) to input-tail, output-tail, and joint-tail regions.

\textbf{Elicitation through Curation} selects improbable or rare inputs from a larger source distribution. CreativeMath, for example, is filtered by human experts from AMC and AIME for problems requiring non-standard or out-of-the-box thinking. 
These problems or prompts by nature are more likely be at the tail region of the task domain distribution. So, compared with the standard benchmark of that domain, creativity benchmarks rise $\text{novelty}(x)$, placing at the \emph{input-tail} (bottom-right of \Cref{fig:elicit-subregions}).

\textbf{Elicitation through Prescription} explicitly requests creativity through a system prompt or task instruction, for example by asking the model to write creatively \citep{chakrabarty2024art} or propose an alternative solution \citep{ye2025assessing}. The input is largely unchanged, but the prompt pushes outputs away from the default, placing the benchmark in the \emph{output-tail} (top-left of \Cref{fig:elicit-subregions}).

\textbf{Elicitation through Constraint} adds restrictions that steer generation away from common outputs, for example by requiring a CS4 story to use a coastal-town setting and flashbacks to the protagonists' youth \citep{atmakuru2024cs4}, or forbidding a particular algorithm in code generation \citep{lu-etal-2025-benchmarking}.
Such combinations can make both prompts and acceptable outputs less common under the reference distribution, shifting evaluation toward the \emph{joint-tail} (top-right of \Cref{fig:elicit-subregions}).


\section{Tailness Contributes to More Accurate Creativity Evaluation}
\label{sec:exp3}

\begin{figure}[!t]
\centering
\includegraphics[width=0.95\linewidth]{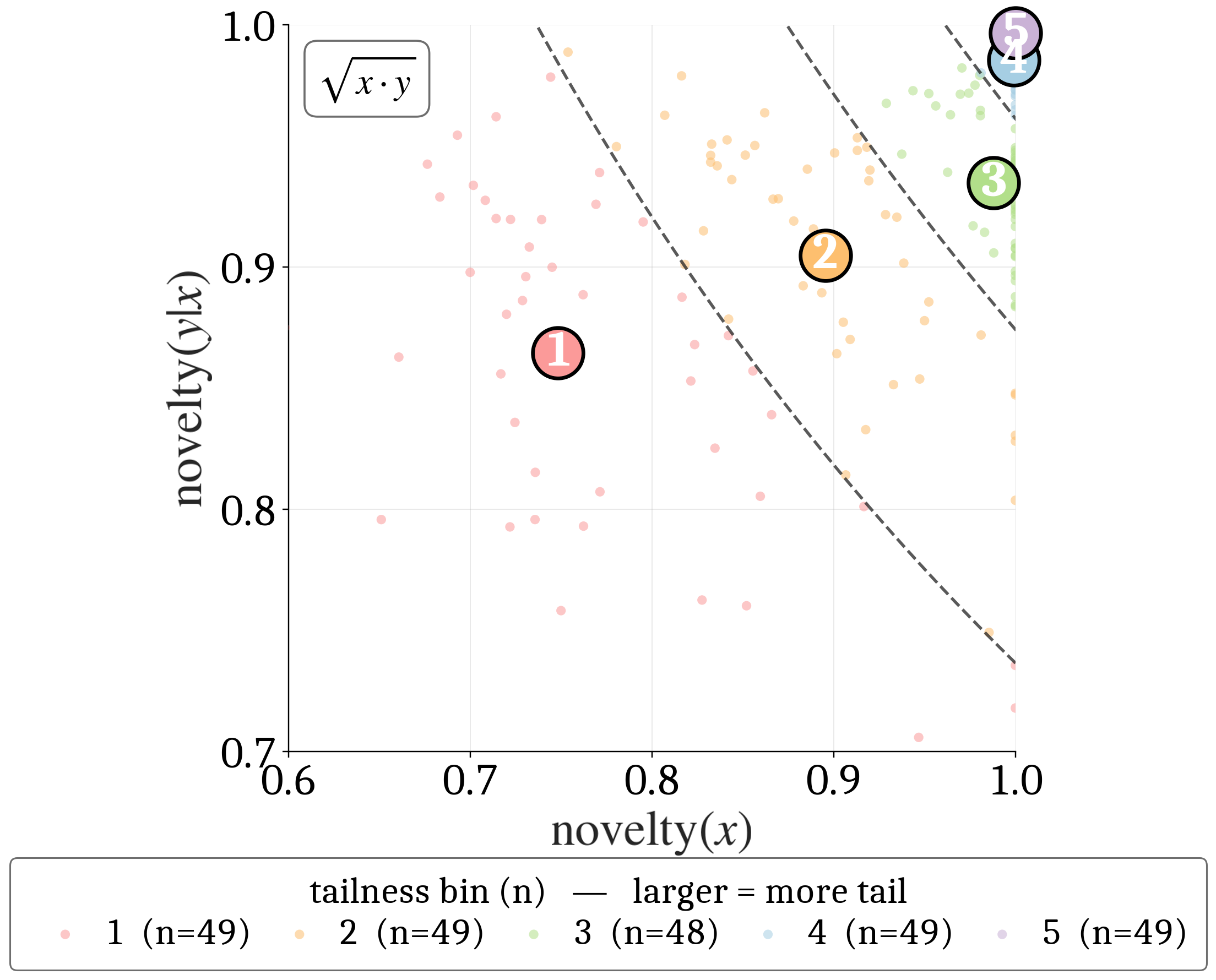}

\vspace{0.8em}
\footnotesize
\setlength{\tabcolsep}{4pt}
\begin{tabular}{c c c c c}
\toprule
\# Bin~~~~ & Input & Output & Joint & Agreement \\ 
& $n(x)$ & $n(y\mid x)$ & $n_{\text{geo}}(x,y)$ & $\tau$ with human\\
\midrule
Bin 1 & 0.748 & 0.864 & 0.799 & $+$0.152 \\
Bin 2 & 0.895 & 0.905 & 0.899 & $+$0.091 \\
Bin 3 & 0.988 & 0.935 & 0.961 & $+$0.061 \\
Bin 4 & 0.999 & 0.985 & 0.992 & $+$\textbf{0.394} \\
Bin 5 & 1.000 & 0.997 & 0.998 & $+$\underline{0.333} \\
\bottomrule
\end{tabular}
\caption{\textbf{Within-benchmark tailness predicts agreement with human creativity rankings} on creativity benchmark (Arena-Hard Creative Writing).
All input-output pairs ($N{=}244$) are split into quintile bins by joint n-novelty under geometric-mean aggregation at $n{=}4$. Bin 1 is the most-peak (lowest-novelty) subset, bin 5 the most-tail.
Within each bin, models are ranked by performance (reward score) and compared to human ranking (LM Arena ELO) via Kendall's $\tau$.
\emph{Top:} scatter of input-output novelty per prompt, colored by bin, with centroids marked.
\emph{Bottom:} per-bin statistics.
Tail bins (4-5) reach $\tau = +0.39, +0.33$, well above peak bins (1-3, $+0.06$ to $+0.15$).
The trend generally holds across other benchmarks and choice of joint-novelty aggregator; see \Cref{app:result}.}
\label{fig:tailness-geo-arena-hard}
\end{figure}

\begin{figure}[!t]
\centering
\includegraphics[width=0.95\linewidth]{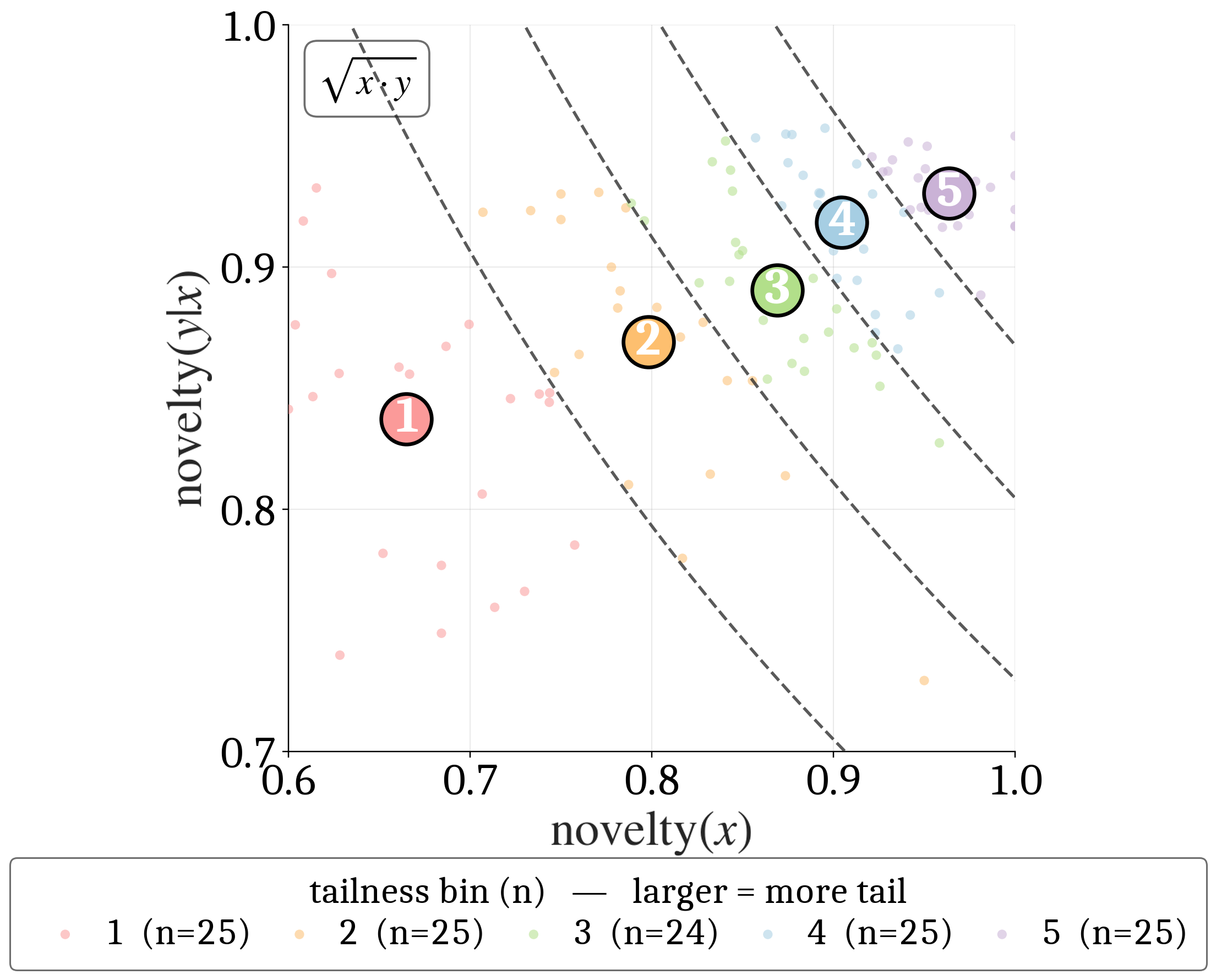}

\vspace{0.8em}
\footnotesize
\setlength{\tabcolsep}{4pt}
\begin{tabular}{c c c c c}
\toprule
\# Bin~~~~ & Input & Output & Joint & Agreement \\ 
       & $n(x)$ & $n(y\mid x)$ & $n_{\text{geo}}(x,y)$ & $\tau$ with human\\
\midrule
Bin 1 & 0.665 & 0.837 & 0.745 & $+$0.099 \\
Bin 2 & 0.798 & 0.869 & 0.832 & $+$0.154 \\
Bin 3 & 0.869 & 0.890 & 0.879 & $+$0.175 \\
Bin 4 & 0.905 & 0.918 & 0.911 & $+$\textbf{0.396} \\
Bin 5 & 0.964 & 0.930 & 0.947 & $+$\underline{0.222} \\
\bottomrule
\end{tabular}

\caption{\textbf{The same trend holds on standard benchmark} (e.g., GPQA-Diamond). 
All input-output pairs ($N{=}124$) are split into quintile bins by joint n-novelty under geometric-mean aggregation, following \Cref{fig:tailness-geo-arena-hard}; Models are ranked by performance (pass@1). 
Tail bins (4-5) reach $\tau = +0.40, +0.22$, above peak bins (1-3, $+0.10$ to $+0.18$).
See other model and joint-novelty in \Cref{app:result}.}
\label{fig:tailness-geo-gpqa}
\end{figure}

\Cref{sec:exp1} shows that creativity benchmarks concentrate on tail regions of $P_\text{ref}$. 
This characterizes how benchmarks are \emph{constructed}, but not whether tail performance is what humans actually reward as creativity. 
We test this first within each benchmark
(\S\ref{sec:tail-within}) and then across writing benchmarks
(\S\ref{sec:tail-across}). 
We use the LM Arena ELO ranking from the creative-writing category\footnote{LM Arena aggregates user-submitted
pairwise preferences between model outputs and converts them to
relative model scores via online Elo updates~\citep{arena_leaderboard_2026}.
Per-category leaderboards are produced by classifying each submitted prompt into topic categories.} as the human reference and Kendall's $\tau$ as the agreement statistic.



\subsection{Within each benchmark, tailness predicts agreement with humans}
\label{sec:tail-within}

\paragraph{Setup.}
For each benchmark $B$, we sort input-output pairs by the shared score $S_B(x)$ defined in \Cref{sec:settings}, keeping pairs with the same input in the same bin.
For visualization, we divide the examples into five equal-sized bins, from
bin~1 (peak, lowest novelty) to bin~5 (most tail). Within each bin, we
rank models by their mean task performance $q$ and compute Kendall's
$\tau_i(B)$ against the LM Arena creative-writing ranking.
We include additional results on other aggregation methods and more bins in \Cref{app:benchmarks}.

\paragraph{Results.}
\textbf{Higher-tail subsets align more closely with the human
creativity ranking.}
\Cref{fig:tailness-geo-arena-hard} shows the pattern on Arena-Hard
Creative Writing: tail bins~4-5 reach $\tau=+0.33$,
compared with $+0.06$ to $+0.15$ for peak bins~1-3. The same pattern
appears on GPQA-Diamond (\Cref{fig:tailness-geo-gpqa}), which is not a benchmark designed specifically for creativity.
This still holds at ten-bin resolution, where the association between bin-level novelty and
human-ranking agreement is statistically significant on both
benchmarks (Spearman $\rho=0.661$-$0.855$, $p=0.0016$-$0.0376$).

\paragraph{Takeaway.} The observed alignment on both creativity and standard benchmarks (\Cref{app:result}) motivates using quality on an existing benchmark's tail as a creativity diagnostic, without redesigning the benchmark or collecting new prompts.\footnote{One additional finding is that the relationship need not be strictly
monotonic at the extreme tail: in the two main examples, agreement
peaks before the final bin. 
Across the broader results in
\Cref{app:result}, however, agreement only occasionally peaks before
the final bin, providing no evidence of a systematic cutoff.}

We denote this diagnostic $q@\mathrm{Tail}(\alpha)$: a model's mean task quality over its input-output pairs in the benchmark's top-$\alpha$ tail, selected using the shared score $S_B(x)$.
It estimates the tail-conditioned quality $A_{\pi,B}(Q_{B,\alpha})$ defined in \Cref{sec:theorm}, with inputs drawn from the benchmark input distribution $\mu_B^X$. Its interpretation and computational requirements are discussed in \Cref{app:q-tail-interpretation}.

\subsection{Across writing benchmarks, the association is positive}
\label{sec:tail-across}

\paragraph{Setup.}
For each writing benchmark $B$, we measure tailness as the mean shared score $S_B(x)$ across its inputs.
We rank models
by their mean quality score $q$ and compute Kendall's $\tau$ between
the resulting ranking and the LM Arena creative-writing ranking.

\paragraph{Results.}
\Cref{fig:tailness-geo} reports the results. Specifically, Arena-Hard Creative
Writing, the most-tail benchmark, reaches $\tau=0.37$ ($p=0.029$),
whereas Alpaca, near the low-novelty end, reaches $\tau=0.16$ and does
not significantly agree with the human ranking ($p=0.37$).
%
Across the eight writing benchmarks, we observe a positive but weaker association (Spearman $\rho=0.26$).\footnote{We understand this as benchmarks differ simultaneously in terms of prompt source, length, format, scoring behavior, sample size, and overall difficulty, which may affect their model rankings independently of tailness. 
This heterogeneity motivates placing greater weight on the controlled
within-benchmark analysis above.}

\paragraph{Interpretation and scope.}
Taken together, the within-benchmark analysis provides the stronger
evidence: higher-tail subsets align significantly better with the
human creativity ranking when benchmark-level factors are held fixed.
Across writing benchmarks, we also observe a positive but weaker association.
We therefore
interpret tailness as a creativity-relevant indicator rather than a
complete characterization of creativity.

Notably, our analysis does not aim to isolate creativity from general
capability. Cleanly separating the two is conceptually difficult,
since creativity requires both novelty and quality, and quality partly
depends on general model capability. Empirically, the LM Arena
creative-writing ranking is itself strongly coupled with the overall
model ranking (\Cref{app:general-capability}). We therefore interpret
the observed alignment as evidence that tail-conditioned performance
captures a creativity-relevant signal, rather than a capability-orthogonal factor.

\begin{table}[t]
\centering
\resizebox{\linewidth}{!}{%
\begin{tabular}{lrrrr}
\toprule
Method
& \makecell{Quality \\ Overall}
& \makecell{Quality \\ at Tail ($q@\text{Tail}$)}
& \makecell{Novelty \\ Overall}
& \makecell{Novelty \\ at Peak} \\
\midrule
Min-$p$ ($p=0.05$)
    & 0.565\textsubscript{\textcolor{increase}{+11.7\%}}
    & 0.558\textsubscript{\textcolor{increase}{+17.7\%}}
    & 0.374\textsubscript{\textcolor{decrease}{-16.3\%}}
    & 0.159\textsubscript{\textcolor{decrease}{-44.2\%}} \\
\textbf{Vanilla Model}
    & \textbf{0.506}\textsubscript{~~~~~~~0\%}
    & \textbf{0.474}\textsubscript{~~~~~~~0\%}
    & \textbf{0.447}\textsubscript{~~~~~~~0\%}
    & \textbf{0.285}\textsubscript{~~~~~~~0\%} \\
Convergent-Divergent Thinking
    & 0.481\textsubscript{\textcolor{decrease}{-4.96\%}}
    & 0.442\textsubscript{\textcolor{decrease}{-6.94\%}}
    & 0.517\textsubscript{\textcolor{increase}{+15.5\%}}
    & 0.322\textsubscript{\textcolor{increase}{+12.9\%}} \\
Temperature Scaling ($\tau \times 1.5$)
    & 0.428\textsubscript{\textcolor{decrease}{-15.4\%}}
    & 0.441\textsubscript{\textcolor{decrease}{-6.96\%}}
    & 0.495\textsubscript{\textcolor{increase}{+10.7\%}}
    & 0.305\textsubscript{\textcolor{increase}{+7.02\%}} \\
Persona
    & 0.442\textsubscript{\textcolor{decrease}{-12.6\%}}
    & 0.421\textsubscript{\textcolor{decrease}{-11.2\%}}
    & 0.484\textsubscript{\textcolor{increase}{+8.28\%}}
    & 0.447\textsubscript{\textcolor{increase}{+56.8\%}} \\
Branch-Solve-Merge
    & 0.438\textsubscript{\textcolor{decrease}{-13.4\%}}
    & 0.408\textsubscript{\textcolor{decrease}{-13.9\%}}
    & 0.508\textsubscript{\textcolor{increase}{+13.6\%}}
    & 0.566\textsubscript{\textcolor{increase}{+98.6\%}} \\
Self-consistency
    & 0.403\textsubscript{\textcolor{decrease}{-20.4\%}}
    & 0.384\textsubscript{\textcolor{decrease}{-19.0\%}}
    & 0.528\textsubscript{\textcolor{increase}{+18.1\%}}
    & 0.654\textsubscript{\textcolor{increase}{+129\%}} \\
Meta-prompting
    & 0.387\textsubscript{\textcolor{decrease}{-23.5\%}}
    & 0.320\textsubscript{\textcolor{decrease}{-32.5\%}}
    & 0.396\textsubscript{\textcolor{decrease}{-11.4\%}}
    & 0.272\textsubscript{\textcolor{decrease}{-4.56\%}} \\
Temperature Scaling ($\tau \times 2.0$)
    & 0.222\textsubscript{\textcolor{decrease}{-56.1\%}}
    & 0.212\textsubscript{\textcolor{decrease}{-55.3\%}}
    & 0.610\textsubscript{\textcolor{increase}{+36.5\%}}
    & 0.904\textsubscript{\textcolor{increase}{+217\%}} \\
CRPO-SFT
    & 0.135\textsubscript{\textcolor{decrease}{-73.3\%}}
    & 0.130\textsubscript{\textcolor{decrease}{-72.6\%}}
    & 0.543\textsubscript{\textcolor{increase}{+21.5\%}}
    & 0.701\textsubscript{\textcolor{increase}{+146\%}} \\
CRPO-DPO
    & 0.097\textsubscript{\textcolor{decrease}{-80.8\%}}
    & 0.098\textsubscript{\textcolor{decrease}{-79.3\%}}
    & 0.560\textsubscript{\textcolor{increase}{+25.3\%}}
    & 0.759\textsubscript{\textcolor{increase}{+166\%}} \\
\bottomrule
\end{tabular}
}
\caption{\textbf{Diagnosing creativity-promoting methods.}
We report overall quality and novelty, together with Quality at Tail ($q@\text{Tail}(\alpha{=}25\%)$) and Novelty at Peak.
Among the evaluated methods, only min-$p$ sampling improves tail quality over the vanilla model, but it reduces both novelty metrics. 
Most methods increase peak novelty but have lower tail quality than the vanilla model.
Results are averaged across 4 representative models and benchmarks.
Subscripts indicate relative percentage changes from the vanilla models; green denotes increases and red denotes decreases. Bold marks the vanilla model.
Full details are provided in \Cref{app:diagnose}.}

\label{tab:diagnose-regions}
\end{table}

\section{Diagnosing Creativity-Promoting Methods}
\label{sec:diagnose}

Our framework motivates examining whether creativity\hyp{}promoting methods improve quality in low-probability regions of $P_{\mathrm{ref}}$.
We use $q@\mathrm{Tail}(\alpha)$ (\Cref{sec:tail-within}) to assess this aspect of their behavior on existing benchmarks.

\paragraph{Setup.}
We apply this diagnostic to a diverse set of creativity-promoting methods,
including test-time interventions (e.g., min-$p$ sampling \citep{nguyen2025turningheatminpsampling}, temperature scaling, persona \citep{xu2025expertpromptinginstructinglargelanguage, pai2025billysteeringlargelanguage}, Convergent-Divergent Thinking \citep{tian-etal-2024-macgyver}, Branch-Solve-Merge \citep{saha2024branchsolvemergeimproveslargelanguage}, Self-consistency \citep{wang2023selfconsistencyimproveschainthought}, Meta-prompting \citep{suzgun2024metapromptingenhancinglanguagemodels}) and training-time methods (e.g., CrPO-SFT and -DPO \citep{ismayilzada2025creativepreferenceoptimization}).
We confirm that both quality and novelty are mentioned as the target aspects for the tasks in prompts.
We include detailed creativity instructions or skills built by \citet{chakrabarty2024art,lu2024llmdiscussionenhancingcreativity,gómezrodríguez2023confederacymodelscomprehensiveevaluation} when the method is needed (e.g., meta-prompting). 
We evaluate each method on a mixture of creativity benchmarks introduced in \Cref{sec:settings}, spanning four model families; full experimental details are provided in \Cref{app:diagnose}.

Conceptually, a method can improve overall creativity along two complementary aspects: 
\textit{i}) {Improving quality on tail quadrants.} When the input or output is novel, the primary challenge is not necessarily to make the output still more novel, but to maintain high output quality despite the unfamiliarity. This capability is measured by Quality at Tail, namely $q@\mathrm{Tail}(\alpha)$;
\textit{ii}) {Improving novelty on peak quadrants.}
A method may also elicit more novel outputs for familiar, high-probability inputs. We measure this capability using Novelty at Peak. 
Based on our proposition, we focus more on the former aspects and less on the latter, since we have shown that most creativity benchmarks generally do not concentrate in the peak quadrant.

\paragraph{Results.}
\Cref{tab:diagnose-regions} shows a quality-novelty trade-off among the evaluated creativity-promoting methods.
Most methods increase Novelty at Peak, but have lower Quality at Tail than the vanilla model.
Conversely, min-$p$ sampling improves Quality at Tail by 17.7\%, but reduces Novelty Overall and Novelty at Peak by 16.3\% and 44.2\%, respectively.
Thus, none of these methods improves both tail quality and peak novelty in the aggregate results.
Additional sanity checks (\Cref{app:diagnose}) with other close-sourced frontier models and quality-focused methods show higher Quality at Tail than the vanilla model, indicating possible room for improvement in tail quality.

Overall, this diagnostic highlights a limitation of current creativity-promoting methods: gains in novelty often coincide with lower quality on uncommon inputs or outputs.
The tail-quality gains achieved by min-$p$ sampling and observed in the sanity checks point to an opportunity to develop methods that improve tail quality while preserving novelty as a potential direction for improving human-perceived creativity.

\section{Creativity is not Identical to Difficulty}
\label{sec:vs_difficulty}

\begin{table}[t]
\centering
\small
\setlength{\tabcolsep}{4pt}
\renewcommand{\arraystretch}{1.2}
\begin{tabular}{llccc}
\toprule
\textbf{Dataset} & \textbf{Split} & \textbf{Diff.} & \textbf{Novelty} & \textbf{Overlap} \\
\midrule
\multirow{3}{*}{\shortstack[l]{MATH\\(n=1{,}000)}}
 & difficulty & 5.00 & 0.728 & 1.00 \\
 & novelty    & 2.38 & 0.965 & \textbf{0.07} \\
 & random     & 2.94 & 0.774 & 0.17 \\
\midrule
\multirow{3}{*}{\shortstack[l]{Cambridge\\CEFR (n=300)}}
 & difficulty & 5.00 & 0.677 & 1.00 \\
 & novelty    & 3.62 & 0.835 & \textbf{0.23} \\
 & random     & 2.65 & 0.693 & 0.12 \\
\bottomrule
\end{tabular}
\caption{\textbf{Difficulty and novelty pick largely disjoint inputs.}
Novelty--difficulty overlap is near or below the random baseline on both
datasets. \textit{Diff.}\ and \textit{Novelty} are per-split means
(difficulty rated 1--5 for math, A2$=$1 to C2$=$5 for Cambridge). Both
datasets are class-balanced.}
\label{tab:difficulty_novelty}
\end{table}

Prior work uses long-tail distributions to explain degraded LLM performance or construct harder evaluations \citep{kandpal2023largelanguagemodelsstruggle, razeghi-etal-2022-impact, mallen-etal-2023-trust, naik-etal-2022-adapting, godbole-jia-2023-benchmarking}. Since our framework emphasizes evaluation in tail regions, a natural question is whether selecting tail examples simply recovers difficult examples.
Empirically, we compare \emph{difficulty} and \emph{novelty} splits on MATH and Cambridge CEFR. The overlap remains low (0.07 and 0.23), near random (\Cref{tab:difficulty_novelty}); the same pattern holds under model-based difficulty on AIME and WritingBench.
Conceptually, the two literatures study different axes. Difficulty work asks whether tail inputs are harder for a fixed model, while our hypothesis asks whether stronger tail performance better aligns with human judgments of creativity across models. See \Cref{app:vs_difficulty} for details.



\section{What Can We Learn?}

From an evaluation perspective, our findings and taxonomy (\Cref{sec:taxonomy}) help assess whether a benchmark's design choices match its intended application and audience.
For example, the reference distribution determines what counts as novel and can affect evaluation results: we use the Pile as a broad textual reference for an evaluation aimed at general users as LM Arena, whereas a benchmark serving scientific researchers may benefit from a reference emphasizing literature from the relevant field.
Designers can also examine whether a benchmark covers the tail regions relevant to its intended use.
Where multiple forms of creativity matter, curation, prescription, and constraint, individually or in combination, can help cover uncommon inputs, outputs, and their combinations.
Metric choices such as judge type and granularity likewise shape which aspects of creativity are measured and involve trade-offs in cost and reproducibility.
Specifying these choices and their rationale alongside the benchmark facilitates replication and comparison with subsequent work.

From a model-development perspective, our findings point to improving model performance specifically in tail regions as a potential pathway toward greater human-perceived creativity. 
Contemporaneous work has begun to investigate long-tail learning in post-training \citep{ramasubramanian2026taillikelihoodreinforcementlearning} and pretraining \citep{badhe2026longtailknowledgelargelanguage, pouransari2026pretraininghierarchicalmemoriesseparating}, through targeted data augmentation and algorithmic interventions. 
Whereas decoding or prompting methods can readily increase output rarity at inference time, they do not necessarily improve a model's ability to produce high-quality outputs in rare regimes. 
Strengthening this ability is likely more challenging, but may offer a more substantive path toward improving model creativity.

\section*{Limitations}

This work studies creativity evaluation through the lens of long-tail performance, but our operationalization is not the only possible one. We approximate the reference distribution using n-gram statistics from the Pile. These statistics capture surface-level rarity but do not directly measure semantic, conceptual, or culturally grounded novelty. Other reference distributions, such as human-authored creative corpora, model self-samples, or domain-specific corpora, may lead to different estimates of tailness.
Model rankings and $q@\mathrm{Tail}$ may also depend on the chosen quality estimator, which can exhibit architecture, length, style, and domain biases. Robustness across reward models, LLM judges and human-calibrated quality measures remains under-explored.

Our empirical validation is also limited by the available benchmarks and human-preference references. Although we evaluate across multiple task domains and models, the strongest human-alignment analysis relies on LM Arena creative-writing rankings, which may not fully represent human creativity judgments in mathematics, scientific reasoning, design, or other creative domains. These rankings reflect holistic preferences from self-selected users rather than judgments that isolate creativity. In addition, benchmark-level correlations are noisy, suggesting that tailness is an important factor but not a complete explanation of creativity evaluation.

Our evaluation of creativity-promoting methods is further limited to open models with 7-8B parameters due to resource constraints. Methods with more complex instructions, such as meta-prompting, may behave differently on larger models or frontier closed-source models.

Finally, $q@\mathrm{Tail}$ is intended as a diagnostic approximation
rather than a replacement for carefully designed creativity
benchmarks. 
Based on the LM Arena creative-writing ranking, creativity-related ranking is itself strongly coupled with overall model capability (\Cref{app:general-capability}), so our experiments cannot attribute the predictive signal of tail performance exclusively to creativity.
High tail performance may also reflect general capability,
task-specific skills, domain knowledge, instruction following, or
robustness to rare inputs. We therefore interpret
$q@\mathrm{Tail}$ as surfacing creativity-relevant behavior under a
declared reference distribution and quality function, rather than as
measuring a capability-orthogonal creativity factor.

\section*{Ethical Considerations}

To the best of our knowledge, this work does not introduce direct ethical harms. Our study focuses on evaluating LLM creativity benchmarks and proposing a diagnostic based on long-tail task performance for creativity evaluation.



\bibliography{custom}

\appendix

\renewcommand{\thefigure}{A\arabic{figure}}
\renewcommand{\thetable}{A\arabic{table}}
\setcounter{figure}{0}
\setcounter{table}{0}


\section{Taxonomy}
\label{sec:taxonomy}

To enable analysis of creativity evaluation for large language models (LLMs), we organize evaluation experiments in the literature along multiple dimensions grouped into two high-level classes: Task and Metric.
The taxonomy decomposes evaluation configurations into independently characterizable components, allowing clearer comparison across prior work and more precise positioning of new experiment designs.
It is constructed to cover most forms of creativity experiments that have been proposed or conceptualized in the literature, thereby supporting more comprehensive and systematic empirical testing.
The dimensions are designed to be comprehensive in two respects: (1) they capture distinctions among experiment setups, particularly those that may lead to divergent empirical conclusions; and (2) for each dimension, we define a set of categories that systematically covers existing setups in the literature while maintaining sufficient granularity to distinguish differences.

\subsection{Task}

In this paper, a task refers to a well-defined type of problem or set of instructions that specifies an intended goal, which includes creativity and admits a space of valid outputs.
A task is defined at the level of objective and problem type rather than the level of specific prompt attributes and surface prompt realization. Variations in tone, theme, wording, or formatting are treated as different prompt realizations of the same task, whereas changes in the core objective (e.g., writing versus problem solving) or knowledge domain (e.g., language versus mathematics) constitute different tasks.

\paragraph{Objective.} The objective defines the primary functional goal the model is expected to achieve. It determines the intended form of successful outputs and the criteria under which they are considered acceptable or valuable.

\paragraph{Domain.} The domain specifies the type of knowledge and competencies required to perform the task. It characterizes the general capability space within which valid solutions would be constructed.

\paragraph{Creativity Elicitation Operation.} This dimension characterizes the operation through which creativity is induced in the task. Creativity may be elicited by an objective or instruction that directly demands it, by posing inputs that are themselves rare or unusual such that standard heuristics no longer suffice, or by imposing explicit constraints that block the most typical outputs and push the model away from its default response.

\paragraph{Targeted Level.} The targeted level specifies where creativity is expected to manifest within the output - whether primarily in surface expression, semantic content, or underlying solution strategies and conceptual structures.

\subsection{Metric}

In this paper, a metric is a concrete algorithmic procedure that takes the model output to one prompt from one decoding pass and produces computational quantitative evaluative signals for analysis or comparison. A metric operationalizes evaluation, but it does not define the task itself.
The task-level results are averaged metrics across all prompts within the task.

\paragraph{Type.} This dimension specifies how creativity is operationalized for measurement at a high level. It distinguishes between approaches that treat creativity as a single construct and those that decompose it into theoretically motivated components evaluated separately.

\paragraph{Granularity.} Granularity defines the unit at which evaluation is computed, thereby determining the structural scale at which variation and novelty are assessed.

\paragraph{Reference Source.} The reference source specifies the distribution, knowledge base, or population relative to which outputs are evaluated. It determines the comparative baseline that grounds judgments.

\paragraph{Output.} This dimension specifies which portion of the model's generation is subject to evaluation, namely whether assessment considers only the final output or also includes intermediate artifacts - reasoning traces, plans, drafts, or actions - produced during task execution.

\paragraph{Judge.} The judge specifies the mechanism responsible for producing the evaluative signal.

%

\onecolumn
{
\footnotesize
\renewcommand{\arraystretch}{1.2}
\setlength{\tabcolsep}{4pt}

\begin{longtable}{
    >{\raggedright\arraybackslash}p{2.4cm}
    >{\raggedright\arraybackslash}p{2.9cm}
    >{\raggedright\arraybackslash}p{9.4cm}
}
\caption{\textbf{Taxonomy of Creativity Evaluation}, organized into Task and Metric.}
\label{tab:cces-taxonomy} \\
\toprule
\textbf{Dimension} & \textbf{Category} & \textbf{Description} \\
\midrule
\endfirsthead

\multicolumn{3}{c}{\footnotesize\textit{Table \ref{tab:cces-taxonomy} continued from previous page}} \\
\toprule
\textbf{Dimension} & \textbf{Category} & \textbf{Description} \\
\midrule
\endhead

\midrule
\multicolumn{3}{r}{\footnotesize\textit{Continued on next page}} \\
\endfoot

\bottomrule
\endlastfoot

\multicolumn{3}{c}{\cellcolor{gray!15}\textbf{TASK}} \\
\midrule

Objective
& Writing
& Tasks whose primary objective is to produce textual artifacts.
\textit{Examples:} Story \citep{chakrabarty2024artartificelargelanguage, atmakuru2024cs4, lu2025salieri}, poetry \citep{west2025basemodelsbeataligned}, essay \citep{kim-etal-2025-representation}, and head-to-head comparison against a professional fiction author \citep{Marco_2024}. 
\\

& Problem Solving
& Tasks whose primary objective is to reach a correct or valid solution.
\textit{Examples:} Math \citep{ye2025assessing, han2025creativity, yu2025deepmathcreative}, 
physical / multimodal \citep{tian2025macgyverlargelanguagemodels, bhavya2024creativeproblemsolving}. 
\\

& Ideation
& Tasks whose primary objective is to propose ideas or generate candidate solutions or associations.
\textit{Examples:} Brainstorming \citep{lv-etal-2024-subjective, ruan2025liveideabenchevaluatingllmsdivergent}, association tasks \citep{chen2023probingcreativitylargelanguage, qiu-hu-2025-deep}, TTCT-style divergent-thinking suites \citep{yao2024assessingunderstanding, zhou2024llmsindividualcollective}, 
and analogy generation \citep{czinczoll2022scientificcreativeanalogies}. 
\\
\midrule

Domain
& Commonsense
& Tasks grounded in everyday knowledge and informal reasoning.
\textit{Examples:} Commonsense QA \citep{talmor2019commonsenseqaquestionansweringchallenge}, Physical Commonsense Reasoning \citep{bisk2019piqareasoningphysicalcommonsense}, PlanBench \citep{valmeekam2023planbenchextensiblebenchmarkevaluating}, RiddleSense \citep{lin2021riddlesense}. 
\\

& Language
& Tasks centered on natural language generation or understanding.
\textit{Examples:} SQuAD \citep{rajpurkar2016squad100000questionsmachine}, GLUE \citep{wang2019gluemultitaskbenchmarkanalysis}, WritingBench \citep{wu2025writingbenchcomprehensivebenchmarkgenerative}, Creative Writing v3 \citep{creative-writing-bench-v3}. \\

& Math
& Tasks involving mathematical reasoning, QA, computation, or problem solving.
\textit{Examples:} GSM8K \citep{cobbe2021trainingverifierssolvemath}, MATH \citep{hendrycksmath2021}, AIME \citep{balunovic_srimatharena_2025}, DeepMath-Creative \citep{yu2025deepmathcreative}. 
\\

& Computer (CS)
& Tasks requiring computer science knowledge or programming skills.
\textit{Examples:} HumanEval \citep{chen2021codex}, SWE-Bench \citep{jimenez2024swebenchlanguagemodelsresolve}. \\

& Science
& Tasks requiring scientific explanation, hypothesis generation, or reasoning.
\textit{Examples:} GPQA \citep{rein2024gpqa}, AstaBench \citep{bragg2025astabench}, scientific-analogy probes \citep{czinczoll2022scientificcreativeanalogies}. 
\\

& Synthetic
& Tasks defined in non-natural-language or symbolic environments.
\textit{Examples:} Symbolic language or knowledge tasks \citep{wang2024languagemodelssolvegraph, nagarajan2025rolldicelook, xu2026muslrmultimodalsymboliclogical}. \\

& Others
& Tasks not covered by the above domains.
\textit{Examples:} Game strategy \citep{west2025basemodelsbeataligned, qian2025escapebenchadvancingcreativeintelligence}, simulation / agentic environments \citep{han2024simulbench}, 
or design ideation tasks. \\
\midrule

{Creativity Elicitation Operation}
& Prescription
& Creativity is demanded through the task objective or explicit instruction.
\textit{Examples:} Creative story writing \citep{chakrabarty2024artartificelargelanguage, atmakuru2024cs4, lu2025salieri, west2025basemodelsbeataligned}, brainstorming \citep{ruan2025liveideabenchevaluatingllmsdivergent}, prompts requiring novel solutions \citep{rabeyah2024llmsagreecreativityevaluation, ye2025assessing}, alternative tool use \citep{rabeyah2024llmsagreecreativityevaluation}. \\

& Curation
& Creativity is induced by posing inputs that are themselves rare or unusual, such that standard heuristics are insufficient to produce a valid output and elicit creativity to finish.
\textit{Examples:} Brainteasers \citep{han2025creativity}, riddles \citep{lin2021riddlesense}, 
creative writing premise, puns and wordplays. \\

& Constraint
& Creativity is induced by adding explicit requirements or constraints to the input that eliminate the most typical outputs, eliciting the model to depart from its common output.
\textit{Examples:} Constrained story writing \citep{atmakuru2024cs4}, constrained code generation \citep{lu2025benchmarkinglanguagemodelcreativity}. \\
\midrule

Targeted Level
& Surface (Form)
& The task implicitly or explicitly demands creativity at the lexical or stylistic level.
\textit{Examples:} Stylistic rewriting tasks \citep{gómezrodríguez2023confederacymodelscomprehensiveevaluation, west2025basemodelsbeataligned}, wordplay. \\

& Meaning (Content)
& The task implicitly or explicitly demands creativity at the level of new semantic combinations or content-level variation.
\textit{Examples:} Story writing \citep{gómezrodríguez2023confederacymodelscomprehensiveevaluation, chakrabarty2024artartificelargelanguage, atmakuru2024cs4, lu2025salieri, west2025basemodelsbeataligned}, divergent association tasks \citep{chen2023probingcreativitylargelanguage, qiu-hu-2025-deep}, TTCT-style ideation \citep{yao2024assessingunderstanding}. 
\\

& Strategy (Concept)
& The task implicitly or explicitly demands creativity at the level of novel ideas, concepts, or solution strategies.
\textit{Examples:} Creative problem-solving or algorithm design tasks \citep{tian2025macgyverlargelanguagemodels, ye2025assessing, han2025creativity}, creative simulation / agentic tasks \citep{han2024simulbench}. 
\\

\midrule
\multicolumn{3}{c}{\cellcolor{gray!15}\textbf{METRIC}} \\
\midrule

{Reference Source}
& Self Samples
& Evaluation relative to the model's own generated samples at test time.
\textit{Examples:} Intra-model diversity \citep{jiang2025artificialhivemindopenendedhomogeneity, zhang2025noveltybenchevaluatinglanguagemodels, atmakuru2024cs4}. \\

& Training Corpus
& Evaluation relative to (a proxy of) the model's training distribution.
\textit{Examples:} Web-scale corpus comparison via n-gram or infinigram lookup \citep{lu2025salieri, padmakumar2025measuringllmnoveltyfrontier, lu2025deathnovelty}. 
\\

& Human Single
& Evaluation from an individual human's judgment relative to their own knowledge.
\textit{Examples:} User rating \citep{stiennon2020learning, ouyang2022training}.
\\

& LLM Single
& Evaluation from another LLM relative to its output or judgment on its own knowledge.
\textit{Examples:} LLM-as-a-judge, inter-model diversity metrics \citep{jiang2025artificialhivemindopenendedhomogeneity}. \\

& Human Community
& Evaluation from a population of humans relative to their own knowledge.
\textit{Examples:} Crowd-sourced human evaluation \citep{chhun2022humancriteria, karpinska2021perils}; story evaluation by multiple literary experts \citep{Marco_2024}; group-level TTCT comparisons against human populations \citep{zhou2024llmsindividualcollective}. 
\\

& LLM Community
& Evaluation from a population of other LLMs relative to their output or judgment on their own knowledge.
\textit{Examples:} Multi-model ensembling and multi-agent evaluation \citep{verga2024replacingjudges, chan2024chateval}; multi-LLM critic ensembles for creative writing \citep{bae2024collectivecritics}. 
\\

& Human Historical
& Evaluation relative to historical human artifacts.
\textit{Examples:} This is an imaginary setting; approximations include internet-scale or library-scale databases. \\
\midrule

Output
& Process
& Evaluation considers intermediate steps, plans, drafts, or actions produced during task execution, rather than or in addition to the final output.
\textit{Examples:} Reasoning trace \citep{han2025creativity}, iterative drafting process \citep{atmakuru2024cs4}, separation of direction- vs.\ process-accuracy for mathematical reasoning \citep{yu2025deepmathcreative}. 
\\

& Outcome
& Evaluation considers only the final output of the model.
\textit{Examples:} Final answer \citep{lu2025salieri, tian2025macgyverlargelanguagemodels, zhang2025noveltybenchevaluatinglanguagemodels}, final story. \\
\midrule

Type
& Monolithic
& Creativity is treated as a unitary construct and measured directly.
\textit{Examples:} Direct human or LLM creativity ratings \citep{gómezrodríguez2023confederacymodelscomprehensiveevaluation, kim-etal-2025-representation, chen2025creval}. 
\\

& Compositional
& Creativity is defined through explicit theoretical components (e.g., novelty and value), each operationalized independently. 
\textit{Note}: We only discuss the novelty component in this research and consider the value component out of our scope. In experiments, we apply the most widely used metrics for the value component (e.g., reward for writing, accuracy for reasoning).
\textit{Examples:} Novelty $\times$ quality decompositions \citep{zhang2025noveltybenchevaluatinglanguagemodels, padmakumar2025measuringllmnoveltyfrontier}; TTCT four-factor decomposition (Fluency / Flexibility / Originality / Elaboration) \citep{yao2024assessingunderstanding}; 
convergent $\times$ divergent decomposition for code \citep{lu2025benchmarkinglanguagemodelcreativity}. \\
\midrule

Granularity
& Token
& Evaluation at the level of individual tokens.
\textit{Examples:} Token probability or entropy \citep{holtzman2020curious}; 
token-level attention-window scoring for creativity detection \citep{yang2025claws}. 
\\

& N-gram
& Evaluation over contiguous token sequences.
\textit{Examples:} Distinct-$n$ \citep{li-etal-2016-diversity}; n-gram corpus-novelty / Creativity Index \citep{lu2025salieri}; 
infinigram-based novelty \citep{lu2025deathnovelty}. 
\\

& Sentence
& Evaluation at the sentence level.
\textit{Examples:} Sentence embedding similarity \citep{reimers-gurevych-2019-sentence, tevet-berant-2021-evaluating}. 
\\

& Step
& Evaluation at the level of procedural or reasoning steps, often composed of multiple sentences.
\textit{Examples:} Step-wise entailment evaluation \citep{prasad-etal-2023-receval, golovneva2023roscoe}. 
\\

& Paragraph
& Evaluation over the paragraph level, which often is the entire generation.
\textit{Examples:} Story evaluation \citep{chakrabarty2024artartificelargelanguage, atmakuru2024cs4}. 
\\

\multicolumn{3}{p{14.5cm}}{\footnotesize\textit{Note: Granularity defines the structural unit at which variation is assessed. Well-designed evaluations align this choice with the task's Targeted Level (e.g., token- or n-gram-level metrics for surface creativity, paragraph-level metrics for content or strategy creativity). Misalignment between granularity and targeted level sometimes appears in existing work which may contribute to inconsistent findings across benchmarks.}} \\
\midrule

Judge
& Deterministic
& Rule-based procedures without trainable parameters.
\textit{Examples:} BLEU, ROUGE, distinct-$n$ \citep{papineni-etal-2002-bleu, lin-2004-rouge, li-etal-2016-diversity}. 
\\

& Learned
& Trained models producing evaluative signals.
\textit{Examples:} Reward models \citep{ouyang2022training}; trained embedding-based scorers \citep{zhang2020bertscore}; 
creativity-specific trained scoring models such as R2C and CLAWS \citep{kim-etal-2025-representation, yang2025claws}; 
fine-tuned creativity evaluators \citep{chen2025creval}. 
\\

& LLM-as-a-Judge
& Evaluation performed by prompting an LLM under a judging protocol.
\textit{Examples:} Rubric-based LLM judging \citep{zheng2023judging, liu2023geval}; 
multi-critic LLM ensembles for creative writing \citep{bae2024collectivecritics}. 
\\

& Human
& Evaluation performed by human annotators.
\textit{Examples:} Expert evaluation \citep{chakrabarty2024artartificelargelanguage} or crowd-sourced evaluation \citep{chhun2022humancriteria}. 
\\

\end{longtable}
}
\twocolumn

\section{Detailed Discussion}

\subsection{Creativity is not Identical to Difficulty}
\label{app:vs_difficulty}

A line of work uses the long tail of a reference distribution to
construct harder evaluations or to explain degraded LLM performance.
Entity frequency, term frequency, and entity popularity in the pretraining corpus \citep{kandpal2023largelanguagemodelsstruggle,razeghi-etal-2022-impact,mallen-etal-2023-trust,naik-etal-2022-adapting}, and pretrained-LM likelihood \citep{godbole-jia-2023-benchmarking}, all predict downstream LLM accuracy; e.g., likelihood-based test splits raise relative error rates by 33-93\% on semantic parsing, NLI, and QA \citep{godbole-jia-2023-benchmarking}.
In all of these works,
tail-ness is invoked to predict or explain difficulty. Our hypothesis
frames creativity as performance on the long tail of $P_\text{ref}$ via
the same kind of operationalization, so a natural worry follows: are
creativity evaluations simply rediscovering difficulty under a new name?

\paragraph{Setup.}
We test this directly. For each source benchmark, we construct three
20\% subsets, a \emph{difficulty} split, a \emph{novelty} split
(tailness under $P_\text{ref}$), and a \emph{random} baseline, and
measure pairwise overlap. We use two operationalizations of difficulty
in parallel: \emph{human-annotated} (from expert labels) and
\emph{model-based} (from aggregated model accuracy). For
human-annotated difficulty we use two class-balanced datasets: MATH
\citep{hendrycks2021measuring}, a competition mathematics benchmark
with five difficulty levels (Level~1-5), and the Cambridge CEFR exam
\citep{mullooly2023cambridge}, a reading comprehension corpus spanning
proficiency levels A2-C2. The difficulty split contains all prompts at
the highest human-annotated level.


\paragraph{Results.}
Across both operationalizations, the difficulty-novelty overlap is
near or below the random baseline.
As shown in \Cref{tab:difficulty_novelty}. On MATH, the overlap is 0.07 against
a random baseline of 0.17, which indicates the two splits are nearly disjoint.
And the novelty split is spread across all five difficulty levels (34\% drawn
from Level~1). 
On Cambridge CEFR, the overlap is 0.23 against a random
baseline of 0.12, which is a small positive correlation, but still largely
distinct. 
Similarly, under model-based difficulty, 
the overlap between the difficulty subset (bottom-20\% by the mean model performance) and the novelty subset sits at the random baseline (0.20 and 0.25 on AIME and WritingBench).
And the model's performance on the novelty split is substantially higher (indicating less difficulty) than on the difficulty split.

Note that we do not claim creativity is orthogonal to
task-specific capability. The empirical splits partially overlap rather
than being fully disjoint, and conceptually, researchers can (and do)
build creativity tests on hard tasks and view part of the capability
for solving hard problems as indicating creativity. The aim of this
paper is not to identify what creativity is or to define its boundaries
against other capabilities; it is to surface the shared implicit
pattern in how existing creativity benchmarks operationalize their
target. See more discussion in \Cref{{app:general-capability}}.

\paragraph{The two literatures are conceptually different.}
The difficulty literature focuses on the \emph{input axis}: for a fixed
model, performance on tail inputs is lower than on peak inputs, and
tail-ness is invoked as one candidate cause alongside task complexity or domain knowledge. 
Our hypothesis focuses on the
\emph{model axis}: on a fixed tail input set, the model whose
performance is higher is likely to be rated as more creative by humans.
Both relations can hold simultaneously. Tail prompts can be harder for any individual model, \emph{and} tail performance can still be the region where between-model differences track creative
capability. 
The experiments in this paper concern the second relation.

\subsection{Creativity and General Model Capability}
\label{app:general-capability}

A potential alternative explanation for our results is that tail
performance primarily distinguishes generally capable models, rather
than capturing a creativity-relevant capability. \textbf{We do not assume that
creativity and general capability are orthogonal.} Indeed, the LM Arena
creative-writing ranking is strongly correlated with its overall model
ranking and with each capability-specific ranking we examined
(\Cref{tab:arena-capability-correlation}).

\begin{table}[!t]
\centering
\small
\begin{tabular}{lrr}
\toprule
Creative-writing ranking vs. & Spearman $\rho$ & $p$ \\
\midrule
Overall               & 0.9443 & $3.92\times10^{-9}$ \\
Instruction Following & 0.8927 & $6.29\times10^{-7}$ \\
Math                  & 0.6739 & $2.17\times10^{-3}$ \\
Coding                & 0.6347 & $4.66\times10^{-3}$ \\
\bottomrule
\end{tabular}
\caption{\textbf{LM Arena creative-writing rankings are strongly
coupled with general model capability.} Spearman correlations compare
the creative-writing model ranking with the overall and
capability-specific LM Arena rankings.}
\label{tab:arena-capability-correlation}
\end{table}

This coupling is also consistent with the novelty-quality account in
\Cref{sec:proposition}: creativity requires not only novelty but also
quality, and quality necessarily depends in part on general language,
reasoning, and instruction-following capabilities. A decomposition
that removes all effects of general capability would therefore also
remove part of the quality component of creativity.

Under this interpretation, the GPQA-Diamond result (\Cref{fig:tailness-geo-gpqa}) is not treated as
evidence that GPQA is itself a complete creativity benchmark. Rather,
it supports the metric-level implication that performance on rare
input-output regions can carry a creativity-relevant signal even on a
benchmark without explicit creativity framing. At the same time, the
current experiments do not establish that this signal is specific to
creativity or separable from general capability. We therefore treat
$q@\mathrm{Tail}$ as a diagnostic approximation, rather than as an
estimate of a disentangled latent creativity factor.

\section{Models and Benchmarks}
\label{app:models-benchmarks}

This appendix summarizes the experimental assets used in §5–§8. We detail the evaluated model families (\Cref{tab:models-creative-writing}) and benchmark (\Cref{tab:benchmarks-revised}) used during generation. These details are provided to improve reproducibility and to clarify how comparisons were standardized across diverse standard and creativity benchmarks.  
 
\label{app:models}

\begin{table*}[t]
\centering
\small
\begin{tabular}{lllll}
\toprule
Model Name & Date & Architecture (Size) & \makecell[l]{LM Arena Creative Writing \\ (Rank / Elo)} & Link \\
\midrule
Qwen3.5-27B & Feb 2026 & 28B Dense & \#26 (1367 $\pm$ 28) & \href{https://huggingface.co/Qwen/Qwen3.5-27B}{Hugging Face} \\
Gemma-3-27b-it & Mar 12, 2025 & 27B Dense & \#34 (1349 $\pm$ 8) & \href{https://huggingface.co/google/gemma-3-27b-it}{Hugging Face} \\
Qwen3.5-35B-A3B & Feb 2026 & 36B (3B MoE) & \#39 (1339 $\pm$ 28) & \href{https://huggingface.co/Qwen/Qwen3.5-35B-A3B}{Hugging Face} \\
Gemma-3-12b-it & Mar 12, 2025 & 12B Dense & \#42 (1334 $\pm$ 23) & \href{https://huggingface.co/google/gemma-3-12b-it}{Hugging Face} \\
Qwen3-30B-A3B-Instruct-2507 & Jul 29, 2025 & 30.5B (3.3B MoE) & \#47 (1325 $\pm$ 11) & \href{https://huggingface.co/Qwen/Qwen3-30B-A3B-Instruct-2507}{Hugging Face} \\
Mistral-Small-3.2-24B-Instruct & Jun 20, 2025 & 24B Dense & \#49 (1322 $\pm$ 12) & \href{https://huggingface.co/mistralai/Mistral-Small-3.2-24B-2506}{Hugging Face} \\
Qwen3-32B & Apr 28, 2025 & 32.8B Dense & \#57 (1307 $\pm$ 22) & \href{https://huggingface.co/Qwen/Qwen3-32B}{Hugging Face} \\
Gemma-3n-E4B-it & Jun 26, 2025 & 8B Raw / 4B Eff. & \#62 (1301 $\pm$ 11) & \href{https://huggingface.co/google/gemma-3n-E4B-it}{Hugging Face} \\
QwQ-32B & Mar 5, 2025 & 32B Dense & \#64 (1297 $\pm$ 10) & \href{https://huggingface.co/Qwen/QwQ-32B}{Hugging Face} \\
OLMo-3.1-32B-Instruct & Jan 13, 2026 & 32.2B Dense & \#67 (1289 $\pm$ 15) & \href{https://huggingface.co/allenai/OLMo-3.1-32B-Instruct}{Hugging Face} \\
Qwen3-30B-A3B & Apr 28, 2025 & 30B (3B MoE) & \#69 (1288 $\pm$ 10) & \href{https://huggingface.co/Qwen/Qwen3-30B-A3B}{Hugging Face} \\
Gemma-3-4b-it & Mar 12, 2025 & 4B Dense & \#77 (1277 $\pm$ 22) & \href{https://huggingface.co/google/gemma-3-4b-it}{Hugging Face} \\
OLMo-3-32B-Think & 2025 & 32B Dense & \#81 (1267 $\pm$ 21) & \href{https://huggingface.co/allenai/OLMo-3-32B-Think}{Hugging Face} \\
Nemotron-3-Nano-30B-A3B & Dec 15, 2025 & 31.6B (3.2B MoE) & \#92 (1252 $\pm$ 13) & \href{https://huggingface.co/nvidia/Nemotron-3-Nano-30B-A3B}{Hugging Face} \\
OLMo-3.1-32B-Think & Jan 2026 & 32B Dense & \#94 (1247 $\pm$ 18) & \href{https://huggingface.co/allenai/OLMo-3.1-32B-Instruct}{Hugging Face} \\
GPT-OSS-20B & Aug 5, 2025 & 21B (3.6B MoE) & \#98 (1236 $\pm$ 18) & \href{https://huggingface.co/open-source/gpt-oss-20b}{Hugging Face} \\
Phi-4 & Dec 12, 2024 & 14B Dense & \#105 (1212 $\pm$ 10) & \href{https://huggingface.co/microsoft/Phi-4}{Hugging Face} \\
Llama-3.1-8B-Instruct & Jul 23, 2024 & 8B Dense & \#119 (1179 $\pm$ 8) & \href{https://huggingface.co/meta-llama/Llama-3.1-8B-Instruct}{Hugging Face} \\
Llama-3.1-Nemotron-Nano-8B-v1 & Mar 18, 2025 & 8B Dense & Unranked & \href{https://huggingface.co/nvidia/Llama-3.1-Nemotron-Nano-8B-v1}{Hugging Face} \\
gemma-4-31B-it & Apr 2, 2026 & 30.7B Dense & Unranked & \href{https://huggingface.co/google/gemma-4-31B-it}{Hugging Face} \\
Qwen3-8B & Apr 28, 2025 & 8B Dense & Unranked & \href{https://huggingface.co/Qwen/Qwen3-8B}{Hugging Face} \\
Qwen3-8B-Thinking & Apr 28, 2025 & 8B Dense & Unranked & \href{https://huggingface.co/Qwen/Qwen3-8B}{Hugging Face} \\
Qwen3-30B-A3B-Thinking & Jul 29, 2025 & 30B (3B MoE) & Unranked & \href{https://huggingface.co/Qwen/Qwen3-30B-A3B}{Hugging Face} \\
OLMo-3-7B-Instruct & Nov 20, 2025 & 7B Dense & Unranked & \href{https://huggingface.co/allenai/OLMo-3-7B-Instruct}{Hugging Face} \\
OLMo-3-7B-Think & Nov 20, 2025 & 7B Dense & Unranked & \href{https://huggingface.co/allenai/OLMo-3-7B-Instruct}{Hugging Face} \\
Nanbeige4-3B-Thinking & Dec 6, 2025 & 3B Dense & Unranked & \href{https://huggingface.co/nanbeige/Nanbeige4-3B-Thinking-2511}{Hugging Face} \\
\bottomrule
\end{tabular}
\caption{Open-weight model properties tracking chronological release dates, architectural configurations, and performance profiles (absolute rank position alongside Elo rating estimation) on the LMSYS Chatbot Arena Creative Writing sub-category leaderboard.}
\label{tab:models-creative-writing}
\end{table*}

\label{app:benchmarks}

\begin{table*}[t]
\centering
\small
\begin{tabular}{lllc}
\toprule
Domain & Benchmark & Type & Link \\
\midrule
Writing & Arena-Hard Creative Writing & Rubric &
\href{https://github.com/lmarena/arena-hard-auto/blob/main/data/arena-hard-v2.0/question.jsonl}{GitHub} \\

Writing & CS4 & Rubric &
\href{https://github.com/anirudhlakkaraju/cs4_benchmark}{GitHub} \\



Writing & StoryGen & Rubric &
\href{https://huggingface.co/datasets/llm-aes/doc-storygen-v2}{Hugging Face} \\

Writing & StoryGen-Px & Rubric &
\href{https://huggingface.co/datasets/llm-aes/doc-storygen-v2}{Hugging Face} \\

Writing & LitBench-Test & RefSim &
\href{https://huggingface.co/datasets/SAA-Lab/LitBench-Test-Release}{Hugging Face} \\

Writing & WritingBench-EN100 & Rubric &
\href{https://github.com/X-PLUG/WritingBench}{GitHub} \\

Writing & Writing-Prompts-Curated-1k & Rubric &
\href{https://huggingface.co/datasets/euclaise/writingprompts}{Hugging Face} \\

Writing & Alpaca & Rubric &
\href{https://huggingface.co/datasets/tatsu-lab/alpaca}{Hugging Face} \\

Writing & TinyStories & RefSim &
\href{https://huggingface.co/datasets/roneneldan/TinyStories}{Hugging Face} \\

Writing & BookMIA & RefSim &
\href{https://huggingface.co/datasets/swj0419/BookMIA}{Hugging Face} \\

Math & AIME-All & EM &
\href{https://huggingface.co/datasets/Pandores/aime-1983-2025}{Hugging Face} \\

Math & CreativeMath & EM &
\href{https://github.com/JunyiYe/CreativeMath/blob/main/data/subset.json}{GitHub} \\


QA & CommonsenseQA (FF) & EM &
\href{https://huggingface.co/datasets/tau/commonsense_qa}{Hugging Face} \\




QA & Braingle-Verifiable & EM &
\href{https://huggingface.co/datasets/ChenLiu1996/Brainteaser}{Hugging Face} \\

QA & MacGyver-Creative & Rubric &
\href{https://github.com/allenai/MacGyver}{GitHub} \\
\bottomrule
\end{tabular}
\caption{
Taxonomy of the 15 evaluation benchmarks, grouped by evaluation
domain and scoring strategy. Benchmark names containing subset,
sampling, filtering, or formatting qualifiers denote the exact
evaluation variants used in our experiments; links point to the
corresponding upstream dataset or official repository.
}
\label{tab:benchmarks-revised}
\end{table*}

\label{app:decoding}
 

\section{Diagnosing creativity-promoting methods settings}
\label{app:diagnose}

This appendix gives implementation details
for the diagnostic experiment in \S\ref{sec:diagnose}.

\paragraph{Models.}
We evaluate the promoting methods on top of four representative 7-8B open-weight models:
Llama-3.1-8B-Instruct, Gemma-3n-E4B-it, OLMo-3-7B-Think, and Qwen3-8B. 
The two thinking models enabled thinking by default.

\paragraph{Data and scoring.}
Each (model, method) cell is run on the same random-sampled 237-prompt creative mix drawn from four
datasets: Arena-Hard Creative Writing, LitBench-test, MacGyver-Creative, and
CreativeMath. 
Quality score from Skywork-Reward-V2-Llama-3.1-8B on the three reward-model datasets, with pass@1 on CreativeMath. 
We normalize per-prompt against each model's baseline run and aggregate into $q$@Tail($\alpha{=}0.25$) using the joint $n$-novelty.

\paragraph{Sanity check.}
To assess whether the diagnostic can capture gains in tail quality, we additionally evaluate CoT as a quality-oriented intervention and Claude Opus 4.6 High as a separate model comparison.
CoT achieves a Quality at Tail score of 0.539, compared with 0.474 for the vanilla model, while Claude Opus 4.6 High achieves 0.610.
These comparisons show that higher tail quality is attainable, although they do not establish joint improvements in tail quality and peak novelty.

\section{Additional Results}
\label{app:result}

\subsection{Creativity Benchmark Lies in the Tail}

\paragraph{Detailed Statical Analysis}
\label{app:distribution-tests}
A pooled comparison within the writing domain further supports this shift: both input novelty (Mann-Whitney $p<0.001$; KS $D=0.108$, $p<0.001$) and output novelty (Mann-Whitney $p<0.001$; KS $D=0.186$, $p<0.001$) differ significantly between creativity
($n=3{,}022$) and standard ($n=6{,}175$) examples, as does their joint
distribution (Hotelling's $T^2$, $p<0.001$). 
The overall pattern is stable across $n\in\{4,5,6\}$ and across model families.

\subsection{Within-Benchmark Robustness and Additional Benchmarks}
\label{sec:additional-within-benchmark}

Section~\ref{sec:tail-within} presents the five-bin
results for Arena-Hard Creative Writing and GPQA-Diamond.
Here, we provide two complementary extensions. First, we
increase the bin resolution from five quantile bins to ten
to test whether the observed relationship is an artifact of
a particular discretization. Second, we report the same
within-benchmark analysis on the remaining benchmarks for
which both task performance and LM Arena creative-writing
ratings are available.

Unless otherwise stated, examples are ordered by joint
4-gram novelty and divided into equal-count bins. The conclusion also holds across different metrics, including Creativity Index and $\text{novelty}_{n=5,6}$; the latter serve as supplementary robustness analyses.
Within each bin, models are ranked by
their mean task performance, and agreement with the LM
Arena creative-writing ranking is measured using
Kendall's $\tau$. Bin~1 denotes the most-peak subset and
the highest-numbered bin denotes the most-tail subset.

\subsubsection{Robustness to the Number of Bins}
\label{sec:additional-ten-bin}

\begin{figure*}[t]
    \centering

    \includegraphics[
        width=0.98\textwidth
    ]{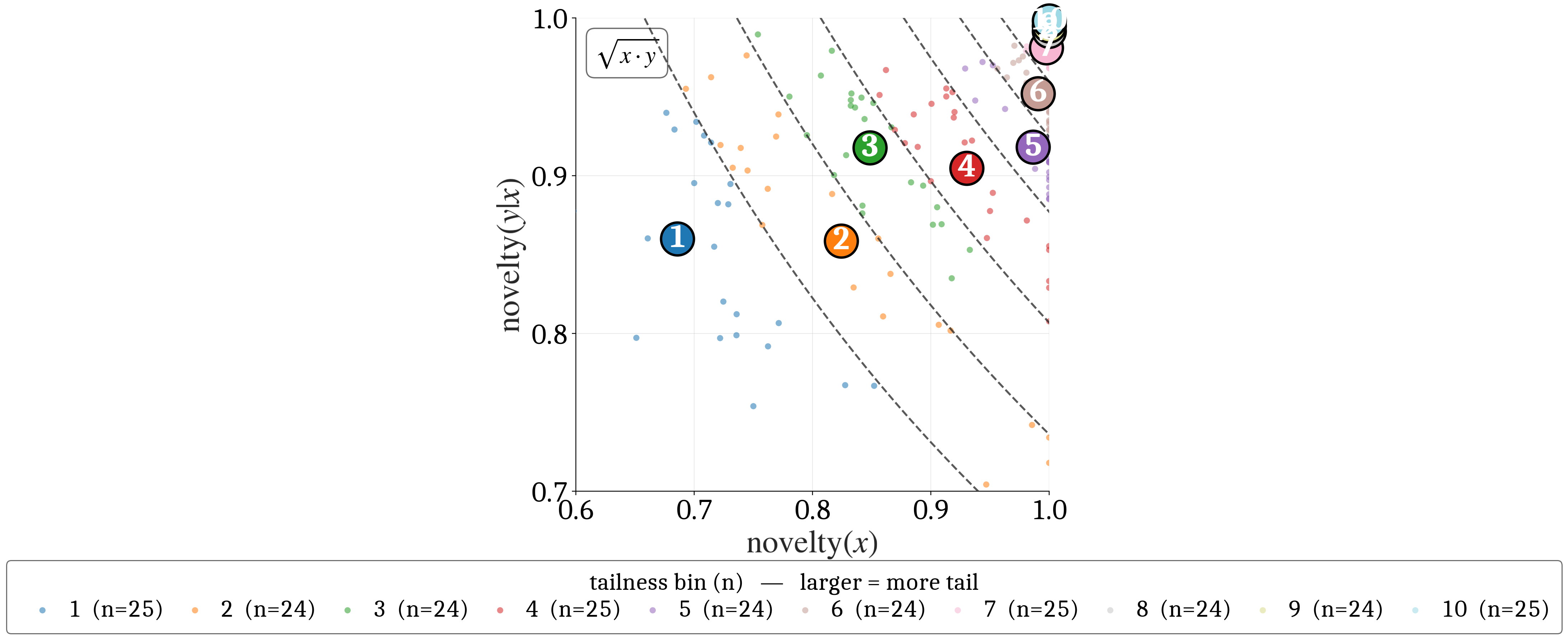}

    \vspace{2mm}
    {\small (a) Arena-Hard Creative Writing}

    \vspace{4mm}

    \includegraphics[
        width=0.98\textwidth
    ]{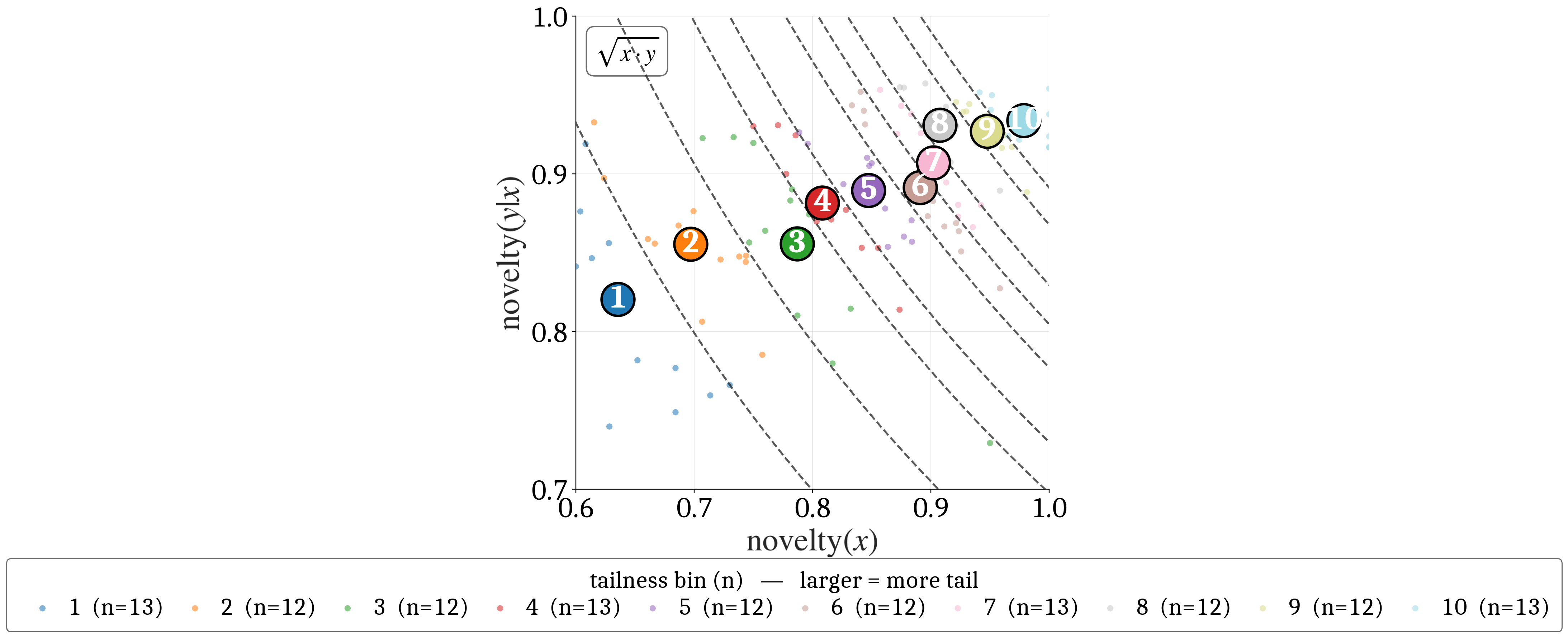}

    \vspace{2mm}
    {\small (b) GPQA-Diamond}

    \caption{
        \textbf{Input-output novelty partitions at ten-bin
        resolution.}
        Input-output pairs are divided into equal-count bins
        using the geometric mean of input and output 4-gram
        novelty. Numbered circles mark bin centroids, dashed
        curves mark equal joint-novelty boundaries, and larger
        bin numbers indicate more-tail examples. The
        corresponding bin-level agreement with the human
        creative-writing ranking is reported in
        Table~\ref{tab:ten-bin-agreement}.
    }
    \label{fig:ten-bin-robustness}
\end{figure*}

Figure~\ref{fig:ten-bin-robustness} shows the ten-bin
partition for Arena-Hard Creative Writing and GPQA-Diamond.
The positive association persists at this finer resolution.
For Arena-Hard, agreement rises from approximately zero in
the earliest bins to $\tau=0.27$-$0.49$ in bins 7--10.
For GPQA-Diamond, bins 7 and 8 reach $\tau=0.43$ and
$\tau=0.45$, respectively, compared with
$\tau=0.06$-$0.20$ for most earlier bins. Across the ten
bins, the Spearman association between bin-level joint
novelty and ranking agreement is positive on both
benchmarks ($\rho=0.661$-$0.855$,
$p=0.0016$-$0.0376$).

The relationship is not strictly monotonic at every adjacent bin. In particular, in some cases, the most-tail bin need not attain the largest agreement. 
As discussed in the main text, extreme-tail subsets can exhibit reduced between-model performance variance,
which weakens the ranking signal even when their novelty is
higher.

\begin{table*}[t]
    \centering
    \scriptsize
    \setlength{\tabcolsep}{3.5pt}
    \begin{tabular}{lrrrrrrrrrrrr}
        \toprule
        Benchmark
        & $\tau_1$
        & $\tau_2$
        & $\tau_3$
        & $\tau_4$
        & $\tau_5$
        & $\tau_6$
        & $\tau_7$
        & $\tau_8$
        & $\tau_9$
        & $\tau_{10}$
        & Mean$_{1:5}$
        & Mean$_{6:10}$ \\
        \midrule
        Arena-Hard
        & 0.000
        & 0.091
        & 0.182
        & 0.000
        & 0.030
        & 0.061
        & 0.273
        & \textbf{0.485}
        & 0.333
        & 0.364
        & 0.061
        & \textbf{0.303} \\

        GPQA-Diamond
        & 0.117
        & 0.064
        & 0.097
        & 0.198
        & 0.085
        & 0.180
        & 0.425
        & \textbf{0.451}
        & 0.178
        & 0.266
        & 0.112
        & \textbf{0.300} \\
        \bottomrule
    \end{tabular}

    \caption{
        \textbf{Ranking agreement at ten-bin resolution.}
        Each entry $\tau_i$ is Kendall's $\tau$ between
        the model ranking induced by performance within
        novelty bin $i$ and the LM Arena creative-writing
        ranking. Bin~1 is the most-peak subset and bin~10
        is the most-tail subset. The last two columns
        compare mean agreement over the peak half
        (bins 1--5) and tail half (bins 6--10).
        Agreement increases substantially in the tail half
        for both Arena-Hard Creative Writing and
        GPQA-Diamond, although the relationship is not
        strictly monotonic across adjacent bins.
    }
    \label{tab:ten-bin-agreement}
\end{table*}

\subsubsection{Results on Additional Benchmarks}
\label{sec:additional-benchmarks}
\begin{figure*}[t]
    \centering

    \begin{minipage}[t]{0.485\textwidth}
        \centering
        \includegraphics[
            width=\linewidth
        ]{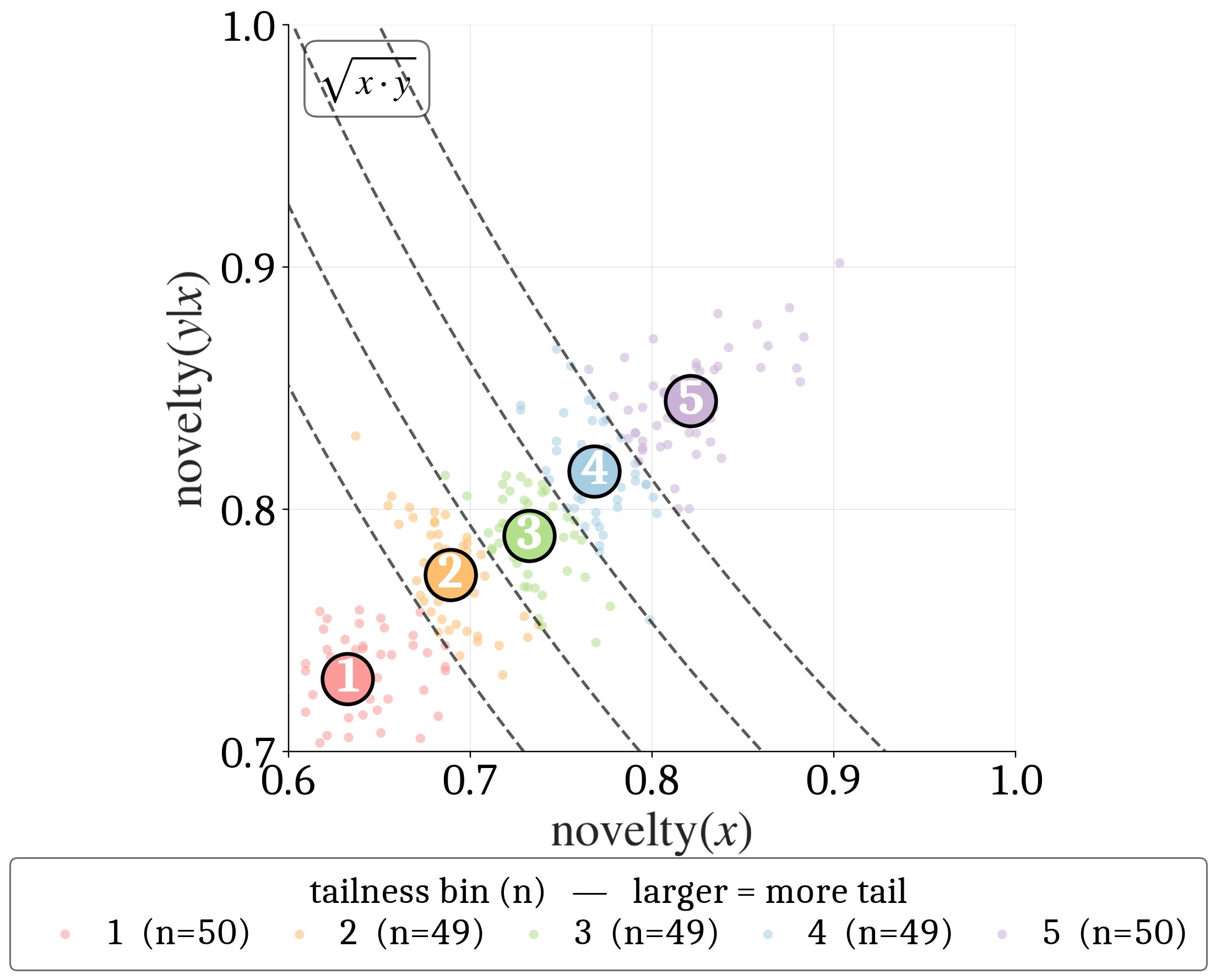}

        {\small (a) BookMIA}
    \end{minipage}
    \hfill
    \begin{minipage}[t]{0.485\textwidth}
        \centering
        \includegraphics[
            width=\linewidth
        ]{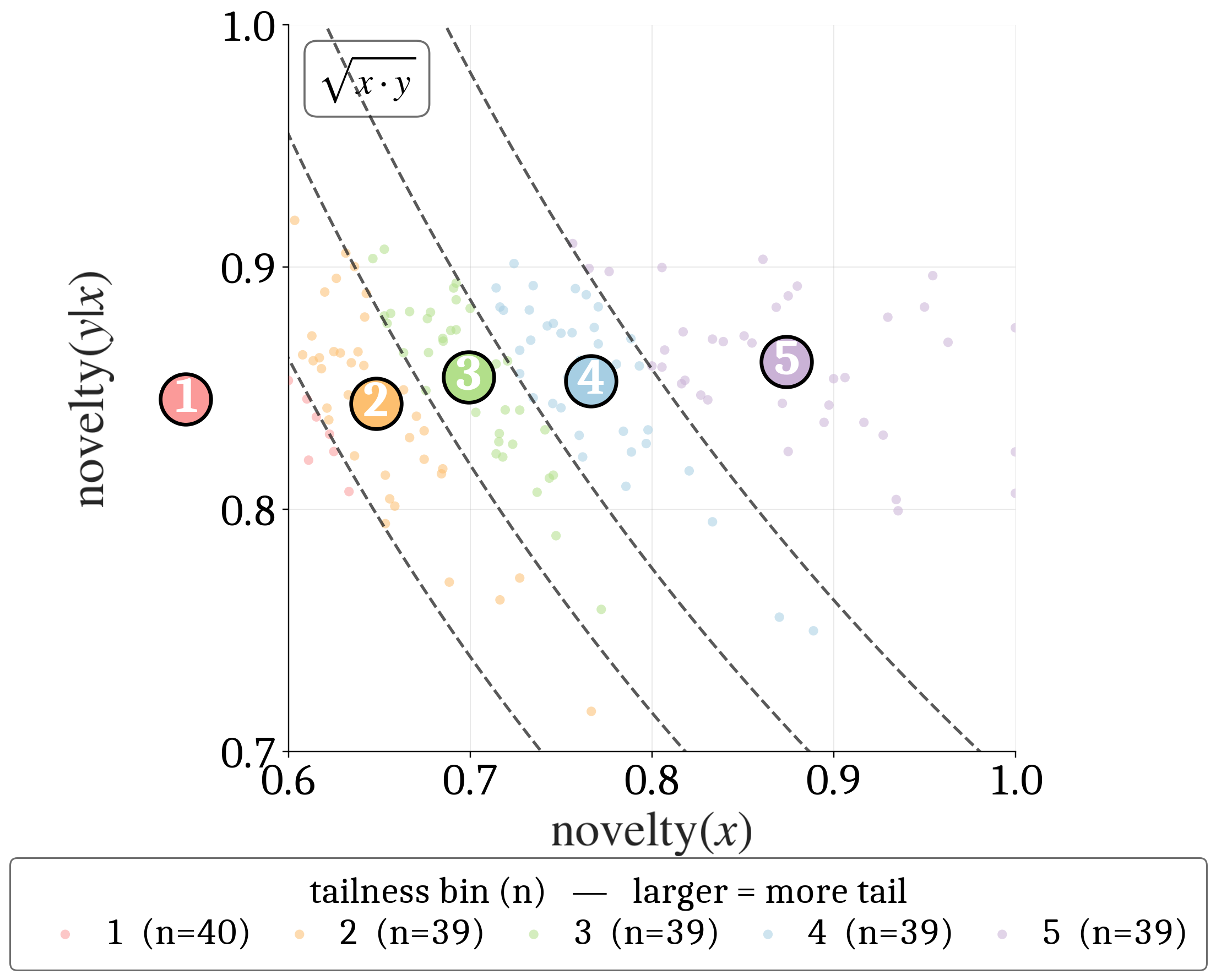}

        {\small (b) StoryGen}
    \end{minipage}

    \vspace{4mm}

    \begin{minipage}[t]{0.485\textwidth}
        \centering
        \includegraphics[
            width=\linewidth
        ]{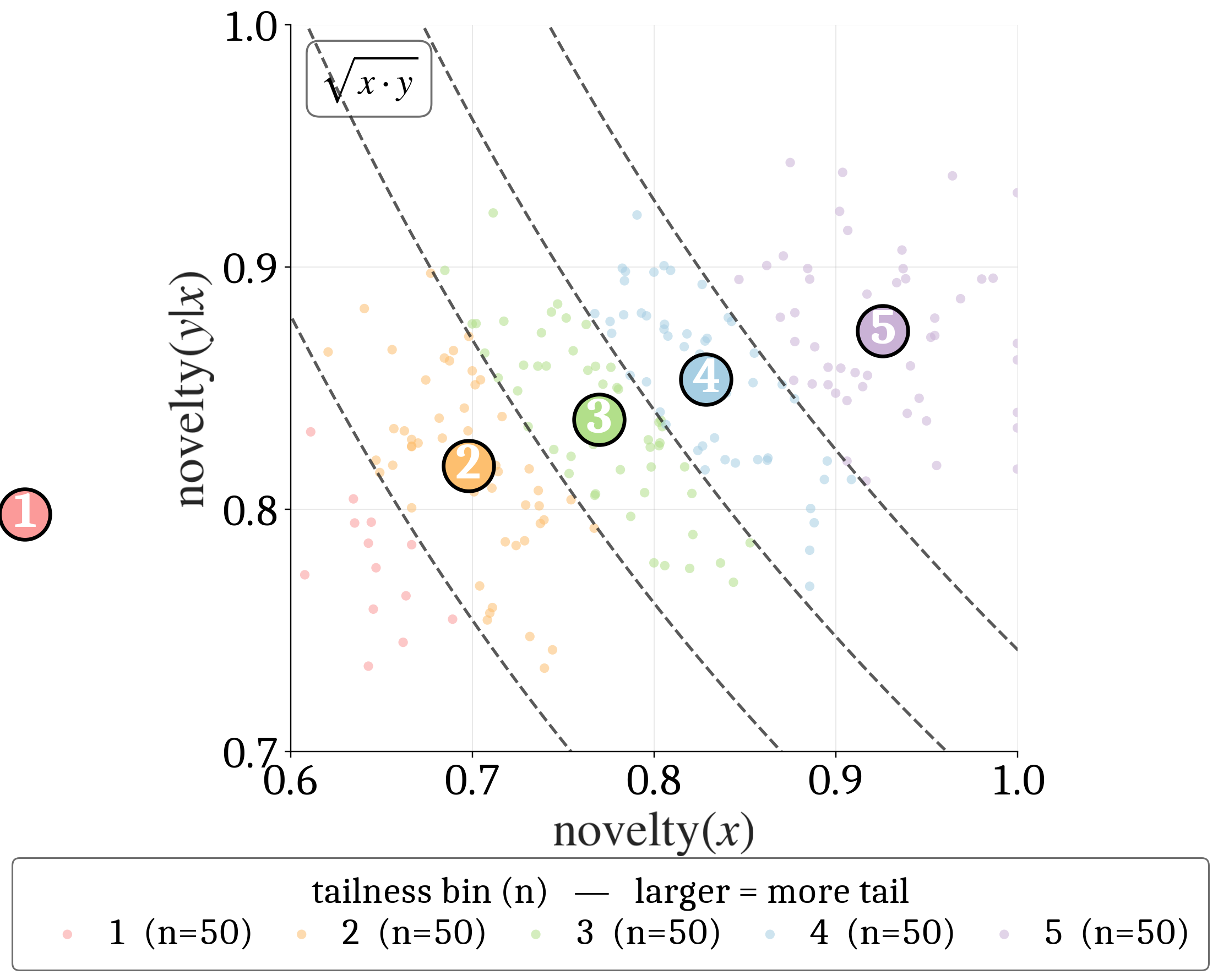}

        {\small (c) AIME-All}
    \end{minipage}
    \hfill
    \begin{minipage}[t]{0.485\textwidth}
        \centering
        \includegraphics[
            width=\linewidth
        ]{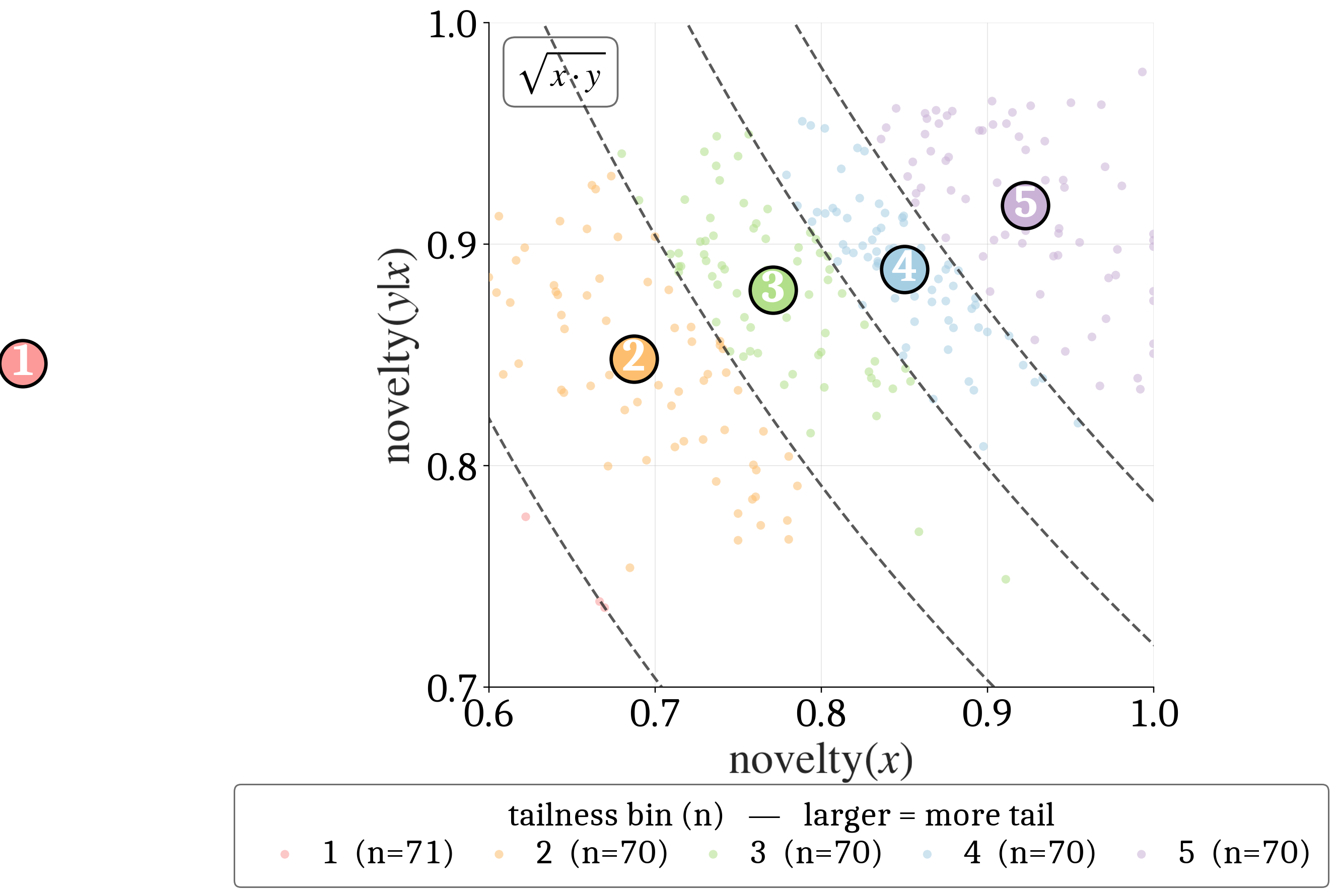}

        {\small (d) Braingle}
    \end{minipage}

    \caption{
        \textbf{Within-benchmark results on additional
        representative benchmarks.}
        BookMIA, AIME-All, and Braingle exhibit higher
        agreement with the LM Arena creative-writing
        ranking in their most-tail bins than in their
        most-peak bins. StoryGen illustrates that the
        relationship need not be monotonic on every
        benchmark. Complete results are reported in
        Table~\ref{tab:additional-bin-agreement}.
    }
    \label{fig:additional-benchmark-results}
\end{figure*}

Figures~\ref{fig:additional-benchmark-results} extend the
five-bin analysis to representative additional benchmarks.
Table~\ref{tab:additional-bin-agreement} reports the
complete bin-level ranking agreement for all benchmarks
included in this analysis.

The results vary across benchmarks, while most of them present non-negative correlation. AIME-All, BookMIA,
and Braingle show higher agreement in their most-tail bins
than in their most-peak bins, consistent with the direction
observed for Arena-Hard and GPQA-Diamond. Alpaca shows
essentially no endpoint difference, and StoryGen and
TinyStories remain positively aligned with the human
creative-writing ranking across all bins. 


These results delimit the scope of the within-benchmark
finding. Tailness is not a universally monotonic predictor
of agreement for every benchmark or task domain. The
relationship is strongest when the quality function and the
evaluated tail region reflect capabilities represented in the
human reference ranking. Domain mismatch, limited
between-model performance variation within individual
bins, and differences in benchmark scoring can weaken or
reverse the observed association. We therefore interpret
tail-conditioned performance as a creativity-relevant
indicator under a domain-aligned reference ranking, rather than
as a domain-invariant measure of creativity.

\begin{table*}[t]
    \centering
    \small
    \setlength{\tabcolsep}{8pt}
    \begin{tabular}{lrrrrrr}
        \toprule
        Benchmark
        & Bin 1
        & Bin 2
        & Bin 3
        & Bin 4
        & Bin 5
        & $\Delta\tau_{5-1}$ \\
        \midrule
        AIME-All
        & 0.131 & 0.095 & 0.135 & 0.089 & 0.290
        & $+0.159$ \\

        Alpaca
        & 0.091 & 0.182 & 0.030 & 0.091 & 0.091
        & $\phantom{+}0.000$ \\

        BookMIA
        & 0.273 & 0.182 & 0.273 & 0.333 & 0.364
        & $+0.091$ \\

        Braingle
        & 0.155 & 0.243 & 0.231 & 0.137 & 0.290
        & $+0.135$ \\


        StoryGen
        & 0.364 & 0.455 & 0.333 & 0.333 & 0.273
        & $-0.091$ \\

        TinyStories
        & 0.303 & 0.212 & 0.273 & 0.303 & 0.212
        & $-0.091$ \\
        \bottomrule
    \end{tabular}

    \caption{
        \textbf{Ranking agreement across joint-novelty
        bins on additional benchmarks.}
        Each entry is Kendall's $\tau$ between the
        performance-induced model ranking within a bin
        and the LM Arena creative-writing ranking.
        Bins are formed using geometric-mean joint
        4-gram novelty; bin~1 is most peak and bin~5 is
        most tail. $\Delta\tau_{5-1}$ is the endpoint
        difference between the most-tail and most-peak
        bins. Positive and null results are reported
        together to avoid selecting benchmarks based on
        the direction of the observed effect.
    }
    \label{tab:additional-bin-agreement}
\end{table*}

\begin{table}[!t]
\centering
\small
\begin{tabular}{lrrrr}
\toprule
Benchmark & \(N\) & Tailness & \(\tau\) vs Arena & \(p_{\mathrm{Kendall}}\) \\
\midrule
Arena-Hard   & 250 & 0.646 & 0.368$^{*}$ & 0.029 \\
CS4-mid      &  50 & 0.357 & 0.286       & 0.093 \\
CS4-light    &  50 & 0.310 & 0.357$^{*}$ & 0.034 \\
BookMIA      & 247 & 0.274 & 0.497$^{*}$ & 0.002 \\
StoryGen     & 200 & 0.255 & 0.485$^{*}$ & 0.003 \\
Alpaca       & 384 & 0.244 & 0.158       & 0.368 \\
CS4-standard &  50 & 0.221 & 0.310       & 0.068 \\
TinyStories  & 220 & 0.176 & 0.333$^{*}$ & 0.049 \\
\bottomrule
\end{tabular}
\caption{\textbf{Across writing benchmarks, tailness (mostly) shows a positive association with human-ranking agreement.}
Tailness averages the shared joint 4-gram novelty score across inputs.
For each input, output novelty is first averaged across the 26 evaluated models and then combined with input novelty using the geometric mean.
\(\tau\) is Kendall's \(\tau\) between each
benchmark's model ranking and the LM Arena ELO ranking. Across the eight
benchmarks, the association between tailness and \(\tau\) is positive
but weak and not statistically significant
(Spearman \(\rho=0.26\), \(p=0.53\)).
\(p_{\mathrm{Kendall}}\) is the p-value for testing the null hypothesis of no rank association between each benchmark's model ranking and the LM Arena ranking; $^{*}$ marks \(p_{\mathrm{Kendall}}<0.05\).
These row-level Kendall tests are separate from the across-benchmark Spearman test.}
\label{fig:tailness-geo}
\end{table}

\subsection{Base-Model Perplexity as an Alternative Novelty Proxy}
\label{app:perplexity-check}

As a robustness check, we used base-model perplexity as an alternative
reference. Its aggregate relationship with the LM Arena
creative-writing ranking is near zero and inconsistent in direction
across models and datasets. This result does not establish that
$n$-gram novelty is uniquely correct; it only shows that perplexity
does not provide a stable alternative for the present evaluation.

\section{Interpretation of $q@\mathrm{Tail}$}
\label{app:q-tail-interpretation}

The shared score $S_B(x)$ fixes novelty weights across models for comparison on the same prompts.
This is a benchmark-level operationalization: $C_{\pi,B}$ need not equal the expected novelty-quality product of a model's own responses.
The integral identity holds for this shared-weight score; the empirical analysis evaluates $q@\mathrm{Tail}$ through agreement with human rankings.

$q@\mathrm{Tail}$ follows a filter-then-score procedure. Tail
membership is determined using the novelty of the input and the
concatenated input-output pairs, with output novelty aggregated across
the evaluated models. The quality function $q$ is then applied only to
examples selected by this joint novelty criterion. Output novelty is
therefore incorporated through tail selection, although $q$ itself
measures only quality and does not explicitly compute a
novelty-quality product.

\paragraph{Computation.}
The corpus index is built once and reused through Infini-Gram-Mini.
Applying $q@\mathrm{Tail}$ reuses existing benchmark prompts, but still requires model generation, novelty computation, and quality evaluation.

\end{document}